\pdfoutput=1

\documentclass{article}

     \usepackage[nonatbib, final]{neurips_2020}

\usepackage[utf8]{inputenc} % allow utf-8 input
\usepackage[T1]{fontenc}    % use 8-bit T1 fonts
\usepackage{hyperref}       % hyperlinks
\usepackage{url}            % simple URL typesetting
\usepackage{booktabs}       % professional-quality tables
\usepackage{amsfonts}       % blackboard math symbols
\usepackage{nicefrac}       % compact symbols for 1/2, etc.
\usepackage{microtype}      % microtypography
\usepackage[square,numbers]{natbib}

\usepackage{sty/mcr}
\usepackage{bbm}
\usepackage{subcaption}
\usepackage{sty/makecell}
\usepackage{sty/graphicx}
\graphicspath{ {./img/} }
\usepackage{amssymb}
\usepackage{bm}
\usepackage{algorithm}
\usepackage{algorithmic}
\usepackage{amsmath}
\usepackage{amsthm}
\usepackage[font=footnotesize]{caption}

\newtheorem{lemma}{Lemma}
\newtheorem{definition}{Definition}

\newcommand{\bmx}{\bm{x}}

\newcommand{\bmtau}{\bm{\tau}}

\DeclareMathOperator*{\argmin}{arg\,min}

\title{Computing Valid $p$-value for Optimal Changepoint by Selective Inference using Dynamic Programming }

\author{%
  Vo Nguyen Le Duy\thanks{Equal contribution} \\
  Nagoya Institute of Technology and RIKEN\\
  \texttt{duy.mllab.nit@gmail.com} \\
   \And
  Hiroki Toda$^\ast$ \\
  Nagoya Institute of Technology\\
  \texttt{toda.h.mllab.nit@gmail.com} \\
   \And
  Ryota Sugiyama \\
  Nagoya Institute of Technology\\
  \texttt{sugiyama.r.mllab.nit@gmail.com} \\
   \And
   Ichiro Takeuchi\thanks{Corresponding author} \\
   Nagoya Institute of Technology and RIKEN \\
  \texttt{takeuchi.ichiro@nitech.ac.jp} \\
}

\begin{document}

\maketitle

\begin{abstract}
Although there is a vast body of literature related to methods for detecting change-points (CPs), less attention has been paid to assessing the statistical reliability of the detected CPs. 
In this paper, we introduce a novel method to perform statistical inference on the significance of the CPs, estimated by a Dynamic Programming (DP)-based optimal CP detection algorithm. 
Our main idea is to employ a Selective Inference (SI) approach --- a new statistical inference framework that has recently received a lot of attention --- to compute exact (non-asymptotic) valid $p$-values for the detected optimal CPs.
Although it is well-known that SI has low statistical power because of over-conditioning, we address this drawback by introducing a novel method called parametric DP, which enables SI to be conducted with the minimum amount of conditioning, leading to high statistical power. 
We conduct experiments on both synthetic and real-world datasets, through which we offer evidence that our proposed method is more powerful than existing methods, has decent performance in terms of computational efficiency, and provides good results in many practical applications.
\end{abstract}

\section{Introduction}
Changepoint (CP) detection is a fundamental problem and has been studied in many areas.
The goal of CP detection is to find changes in the underlying mechanism of the observed sequential data.
%
%Analyzing the detected CPs benefits to several applications such as bioinformatics \cite{frick2014multiscale, pierre2014performance}, financial analysis \cite{fryzlewicz2014wild}, climatology \cite{killick2012optimal}, and signal processing \cite{jandhyala2013inference}.
Analyzing the detected CPs benefits to several applications \cite{frick2014multiscale, pierre2014performance, fryzlewicz2014wild, killick2012optimal, jandhyala2013inference}.
There is a vast body of literature related to methods for detecting CPs \cite{auger1989algorithms, vostrikova1981detecting, olshen2004circular, tibshirani2005sparsity, maidstone2017optimal, fearnhead2019detecting, van2020evaluation} --- nice surveys can be found in \cite{aminikhanghahi2017survey, truong2019selective}.
CP detection is usually formulated as the problem of minimizing the cost over segmentations, where \emph{Dynamic Programming (DP)} is commonly used because it can solve the minimization problem efficiently, and exactly find the optimal CPs under the given criteria.

Unfortunately, less attention has been paid to the statistical reliability of the detected CPs. 
Without statistical reliability, the results may contain many \emph{false detections}, i.e., the detected CPs may not be true CPs. 
These falsely detected CPs are harmful when they are used for high-stake decision making such as medical diagnosis or automatic driving.
Therefore, it is highly necessary to develop a \emph{valid statistical inference} for the detected CPs that can properly control the risk of false detection.

Valid statistical inference on CPs is intrinsically difficult because the observed data is used twice --- one for detection and another for inference, which is often referred to as \emph{double dipping}~\cite{kriegeskorte2009circular}. 
In statistics, it has been recognized that naively computing $p$-values in double dipping is highly biased, and correcting this bias is challenging.
Our idea is to introduce \emph{Selective Inference (SI)} framework for resolving this challenge.

\begin{figure}[t]
\centering
\includegraphics[width=0.95\linewidth]{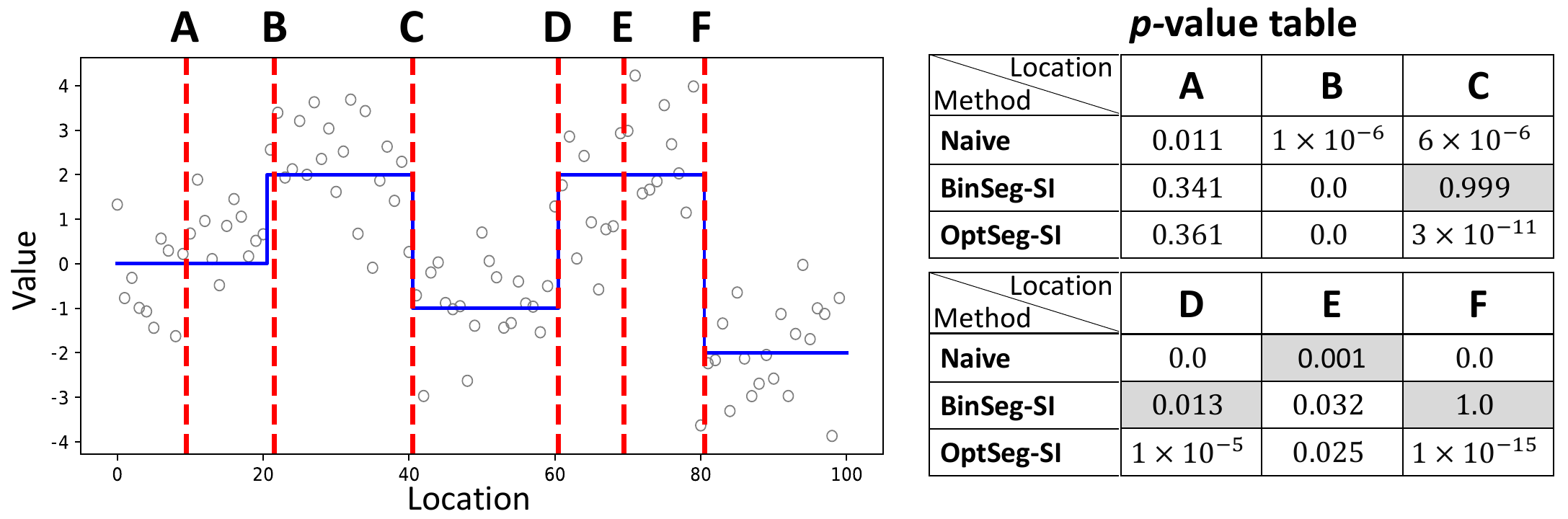}  
\caption{
An illustrative example of the problem and the methods considered in this paper. 
The blue line and the grey circles indicate the underlying mean and the observed sequence, respectively.
The red dotted lines are the results of optimal segmentation (OptSeg) and binary segmentation (BinSeg).
Here, the results of OptSeg and BinSeg were the same.
With Bonferroni correction, to control false detection rate at $0.05$, the significance level is decided by $\frac{0.05}{6} \approx 0.008$.
The naive $p$-value is small even for falsely detected CP ({\bf E}). 
BinSeg $p$-values can identify falsely detected CPs, but it fails to detect some true CPs ({\bf C}, {\bf D}, {\bf F}) due to the lack of power. 
The proposed $p$-values (OptSeg-SI) can successfully identify both true positive and false positive detections. 
}
\label{fig:intro}
\end{figure}

\textbf{Existing works and their drawbacks.} 
In the case of testing for single CP,
most of the existing inference methods rely on \emph{asymptotic} distribution of the maximum discrepancy measure, such as CUSUM score~\cite{page1954continuous}, Fisher discriminant score~\cite{mika1999fisher, harchaoui2009kernel}, and MMD~\cite{li2015m}, which is derived under some restrictive assumptions such as weak dependence among the data points. 
Asymptotic inference for multiple CPs was proposed by \cite{frick2014multiscale} under the name of \emph{SMUCE}.
%\emph{Simultaneous Multiscale Changepoint Estimator (SMUCE)}. 
%
These asymptotic approaches often fail to control type I error when the sequence is short or contains highly correlated data points.
Besides, it has been observed that these approaches are often conservative, i.e., low statistical power~\cite{hyun2018exact}.
% - we also empirically confirm that SMUCE has this property in \S \ref{sec:sec_exp}.

%The asymptotic issue can be addressed by \emph{Selective Inference (SI)}, which allows us to derive the \emph{exact} (non-asymptotic) sampling distribution of the test statistic.
%
In the past few years, SI has been actively studied and applied to various problems~\cite{bachoc2014valid, fithian2015selective, choi2017selecting, tian2018selective, chen2019valid, bachoc2018post, loftus2014significance, loftus2015selective, panigrahi2016bayesian, tibshirani2016exact, lee2016exact, yang2016selective, suzumura2017selective, tanizaki2020computing, duy2020parametric, duy2020quantifying}.
The basic idea of SI is to make inference conditional on the selection event, which allows us to derive the \emph{exact} (non-asymptotic) sampling distribution of test statistic.
% by Lasso. 
%
However, characterizing the necessary and sufficient selection event is computationally challenging.
For example, in \cite{lee2016exact}, the authors considered inference conditional not only on the selected features but also on their signs for computational tractability.
%~\footnote{\cite{liu2018more} has recently developed an approach for SI conditional only on the selected features by LASSO.}.
%
However, such an \emph{over-conditioning} leads to loss of power~\cite{lee2016exact, liu2018more, duy2020parametric}. 

SI was first discussed in the context of CP detection problem by Hyun et al. \cite{hyun2018exact}, in which the authors studied Fused Lasso. 
Later, Umezu et al. \cite{umezu2017selective} and Hyun et al. \cite{hyun2018post} studied SI for CUSUM-based CP detection  and binary segmentation, respectively. 
Unfortunately, these methods inherit the drawback of other SI studies, i.e., the loss of power by over-conditioning. 
In other words, the inference is made not only conditional on the detected CPs, but also on other unnecessary extra events.
\paragraph{Contributions.} We provide an exact (non-asymptotic) inference method for optimal CPs, which we call \emph{OptSeg-SI}, based on the concept of SI. 
To our knowledge, this is the first method that can provide valid $p$-values to the CPs detected by DP.
Unlike existing SI approaches for CPs~\cite{hyun2018exact, hyun2018post, umezu2017selective}, the inference in the OptSeg-SI method is made under the minimum amount of conditioning, leading to high statistical power.
To this end, we develop a new method called \emph{parametric DP}, which enables us to efficiently characterize the selection event.
We conduct experiments on both synthetic and real-world datasets, by which, we offer the evidence that the OptSeg-SI
      1) is more powerful than the existing methods \cite{hyun2018post, frick2014multiscale},
      2) successfully controls false detection probability,
      3) has good performance in terms of computational efficiency,
      and
      4) provides good results in many practical applications. 

Figure \ref{fig:intro} shows an illustrative example of the problem and the methods we consider in this paper.
For reproducibility, our implementation is available at
\begin{center}
\url{https://github.com/vonguyenleduy/parametric_selective_inference_changepoint}
\end{center}

\section{Problem Statement}
We
consider 
CP detection problem
for
mean-shift,
%\footnote{More complicated CP detection problem can often be transformed into mean-shift CP detection problem. 
%%
%For instance, in kernel CP detection \cite{li2015m}, the original signal is  transformed to be piecewise constant, and the objective is then to detect mean-shifts in the transformed signal.}, 
which is the most 
studied model in the
literature,
and has been applied to many real-world applications, especially in bioinformatics \cite{muggeo2011efficient, chen2008statistical}.
Mean-shift CP detection is the base of many other CP detection methods.
If one knows what kind of changes to focus on (e.g., changes in variance), we can convert the problem into mean-shift CP detection. 
Otherwise, nonparametric CP detection methods such as kernel CP detection \cite{li2015m} can be used. It is well known that many nonparametric methods can be cast into a mean-shift CP detection. 
Therefore, mean-shift CP detection is worth investigating as a canonical form of the more complex problems.

Let us consider a random sequence
%\begin{align}
% \label{eq:model}
$
 \bm X = (X_1, \ldots, X_N)^\top \sim \NN(\bm \mu, \bm \Sigma)
$,
%\end{align}
where
$N$
is the length,
% of the sequence,
$\bm \mu \in \RR^N$
is
unknown mean vector,
and
$\bm \Sigma \in \RR^{N \times N}$
is covariance matrix
which is known or estimable from external data~\footnote{The covariance matrix $\bm \Sigma$ is typically estimated by ``null'' sequences which are known to have no CP (see \citet{takeuchi2009potential} for an example in bioinformatics).}.
%
%When there are multiple \emph{true} CPs,
%% in the sequence,
%the mean vector
%$\bm \mu = (\mu_1, \ldots, \mu_N)^\top$
%is \red{\st{represented as a piecewise constant}} \blue {a} sequence
%having shifts at these CPs.
%
Given an observed sequence %sampled from the model \eq{eq:model} as
$
 \bm x^{\rm obs} = (x^{\rm obs}_1, \ldots, x^{\rm obs}_N)^\top \in \RR^N, 
$
the goal of CP detection is to estimate the true CPs. 
The vector of detected CP locations is denoted
%by a CP detection algorithm 
as 
$
 \bm \tau^{\rm det} = (\tau^{\rm det}_1, \ldots, \tau^{\rm det}_K),
$
where
$K$
is the number of %detected 
CPs,
and
$\tau^{\rm det}_1 < \cdots < \tau^{\rm det}_K$
are the 
%$K$ detected 
CP locations 
(%for notational simplicity,
we %additionally 
set
$\tau^{\rm det}_0=0$
and
$\tau^{\rm det}_{K+1}=N$).
%
%Let $s$ and $e$ be a pair of indices which satisfy $1 \le s \le e \le N$. 
%
We define $\bm x_{s:e} \sqsubseteq \bm x$ as a subsequence of $\bm x \in \RR^N$ 
from positions
$s$
to
$e$,
where 
%$s$ and $e$
%satisfy 
$1 \le s \le e \le N$.
%The subsequence of a sequence
%$\bm x \in \RR^N$ 
%from positions
%$s$
%to
%$e$
%is denoted as
%$\bm x_{s:e} \sqsubseteq \bm x$.
%
The average
of 
%the subsequence
$\bm x_{s:e}$ 
is written as 
$\bar{x}_{s:e} = \frac{1}{e-s+1} \sum_{i=s}^e x_i$, 
and the cost function which measures the ``homogeneity" of $\bm x_{s:e}$ is defined as 
$C(\bm x_{s:e}) = \sum_{i=s}^e (x_i - \bar{x}_{s:e})^2$.
%
%\begin{equation}
%	C(\bm x_{s:e}) = \sum_{i=s}^e (x_i - \bar{x}_{s:e})^2.
%\end{equation}
%
\subsection{Optimal CP detection}
Although we do not assume any true structures in the mean vector $\bm \mu = (\mu_1, \ldots, \mu_N)^\top$, we consider the case where data analyst believes that the data can be reasonably approximated by a piecewise constant function.
When the number of change points $K$ is known, it is reasonable to formulate the CP detection problem as the following optimization problem
\begin{align}
 \label{eq:opt1}
 \bm \tau^{\rm det} &= \arg \min_{\bm \tau} \sum_{k=1}^{K+1} C(\bm x^{\rm obs}_{\tau_{k-1}+1: \tau_{k}}).
\end{align}
%
%On the other hand,
When the number of CPs $K$ is unknown,
the CP detection problem is defined as
\begin{align}
 \label{eq:opt2}
 \hspace{-1mm}\bm \tau^{\rm det}
 =
 \arg \min_{\bm \tau} \sum_{k=1}^{{\rm dim}(\bm \tau) +1} C(\bm x^{\rm obs}_{\tau_{k-1}+1:\tau_{k}}) + \beta {\rm dim}(\bm \tau), 
\end{align}
where
${\rm dim}(\bm \tau)$
is the dimension of a CP vector $\bm \tau$, %i.e., the number of CPs, 
and
$\beta \in \RR^+$
is a hyper-parameter,
%representing the degree of penalization, 
which can be defined based on
several methods such as BIC~\cite{schwarz1978estimating}. 
%or AIC~\cite{akaike1974new}.
%
The optimal solutions of %the minimization problems
\eq{eq:opt1}
and
\eq{eq:opt2}
can be %easily 
obtained by %using 
DP.
%~\cite{auger1989algorithms, jackson2005algorithm}.
%
\begin{definition}
We denote the event that
the optimal CP vector
$\bm \tau^{\rm det}$
is
detected by applying DP algorithm $\cA$ to the observed sequence
$\bm x^{\rm obs}$ as
\begin{align}
\bm \tau^{\rm det} = \cA(\bm x^{\rm obs}). 
\end{align}

\end{definition}

\subsection{Inference on the detected CPs}
%
%Now, let us consider statistical inference on the detected CPs
%$\tau^{\rm det}_1, \ldots, \tau^{\rm det}_K$.
%
For the inference on the $k^{\rm th}$ detected CP $\tau^{\rm det}_k$,
$k \in [K]$,
we consider the following statistical test
%
%\begin{align}
% \nonumber
% &
%\hspace{-3mm}{\rm H}_{0,k}: \mu_{\tau^{\rm det}_{k-1}+1} = \cdots = \mu_{\tau^{\rm det}_k} = \mu_{\tau^{\rm det}_k + 1} = \cdots = \mu_{\tau^{\rm det}_{k+1}}
%\\
%\label{eq:hypotheses}
% & \hspace{40mm} {\rm vs.}
% \\
% \nonumber
% &
% \hspace{-3mm}{\rm H}_{1,k}: \mu_{\tau^{\rm det}_{k-1}+1} = \cdots = \mu_{\tau^{\rm det}_k} \neq \mu_{\tau^{\rm det}_k + 1} = \cdots = \mu_{\tau^{\rm det}_{k+1}},
%\end{align}
%
\begin{align}
 \nonumber
 &
{\rm H}_{0,k}: \frac{1}{\tau^{\rm det}_{k}-\tau^{\rm det}_{k-1}}  \left (\mu_{\tau^{\rm det}_{k-1}+1} + \cdots + \mu_{\tau^{\rm det}_k} \right ) = \frac{1}{\tau^{\rm det}_{k+1} - \tau^{\rm det}_k} \left (\mu_{\tau^{\rm det}_k + 1} + \cdots + \mu_{\tau^{\rm det}_{k+1}} \right)
\\
\label{eq:hypotheses}
 & \hspace{65mm} {\rm vs.}
 \\
 \nonumber
 &
{\rm H}_{1,k}:\frac{1}{\tau^{\rm det}_{k}-\tau^{\rm det}_{k-1}}  \left (\mu_{\tau^{\rm det}_{k-1}+1} + \cdots + \mu_{\tau^{\rm det}_k} \right ) \neq \frac{1}{\tau^{\rm det}_{k+1} - \tau^{\rm det}_k} \left (\mu_{\tau^{\rm det}_k + 1} + \cdots + \mu_{\tau^{\rm det}_{k+1}} \right),
\end{align}
where $[K] = \{1, ..., K\}$ indicates the set of natural numbers up to $K$.
A natural choice of the test statistic is the difference between the average of the two segments before and after the $k^{\rm th}$ CP
%, which is written as 
\begin{align}
 \label{eq:test_statistic}
 \bm \eta_k^\top \bm X = \bar{X}_{\tau^{\rm det}_{k-1}+1: \tau^{\rm det}_k} - \bar{X}_{\tau^{\rm det}_k + 1: \tau^{\rm det}_{k+1}},
\end{align}
where
$
\bm \eta_k = \frac{1}{\tau^{\rm det}_{k}-\tau^{\rm det}_{k-1}} \one^N_{\tau^{\rm det}_{k-1}+1:\tau^{\rm det}_k} - \frac{1}{\tau^{\rm det}_{k+1} - \tau^{\rm det}_k} \one^N_{\tau^{\rm det}_k+1:\tau^{\rm det}_{k+1}},
$
and $\one^N_{s:e} \in \RR^N$ is a vector whose elements from position $s$ to $e$ are set to 1, and 0 otherwise.
Remember that we do \emph{not} assume that the true mean values are piecewise constant, i.e., we do not assume that $\mu_{\tau_{k-1}^{\rm det}+1} = \ldots = \mu_{\tau_k^{\rm det}}$ nor $\mu_{\tau_{k}^{\rm det}+1} = \ldots = \mu_{\tau_{k+1}^{\rm det}}$. Even without assuming true piecewise constant functions, the population quantities in \eq{eq:hypotheses} are well-defined as the best constant approximations of the subsequences between two detected CPs.

Suppose, for now, that the hypotheses in \eq{eq:hypotheses} are fixed, i.e., non-random. %then for testing the significance of the $k^{th}$ CP,
Then, the \emph{naive} (two-sided) $p$-value 
%in classical z-test 
%
is given as 
\begin{align}
\label{eq:naive-p}
 \hspace{-2mm} p^{\rm naive}_k
 =
 \PP_{{\rm H}_{0,k}}
 \left(
 |\bm \eta_k^\top \bm X| \ge |\bm \eta_k^\top \bm x^{\rm obs}|
 \right)
 =2 \min \{
 F_{0, \bm \eta_k^\top \Sigma \bm \eta_k}(\bm \eta_k^\top \bm x^{\rm obs}),
 1 - F_{0, \bm \eta_k^\top \Sigma \bm \eta_k}(\bm \eta_k^\top \bm x^{\rm obs})
 \},
\end{align}
where
$
F_{m, s^2}
$
is the c.d.f. of Normal distribution $\NN(m, s^2)$. 

However,
%when the hypotheses in \eq{eq:hypotheses} are selected by the data, 
since the hypotheses in \eq{eq:hypotheses} are actually not fixed in advance,
the naive $p$-value is not \emph{valid} 
in the sense that,
if we reject
${\rm H}_{0, k}$
%when
with a 
%certain 
significance level
$\alpha$
(e.g., $\alpha=0.05$),
the false detection rate (type-I error) cannot be controlled at level $\alpha$.
This is due to the fact that the hypotheses in \eq{eq:hypotheses} 
are \emph{selected} by data,
and \emph{selection bias} exists.
One way to avoid the selection bias 
%problem 
is to consider
%the inference based on 
the sampling distribution of a test statistic
\emph{conditional} on the selection event.
%
%Instead of using naive $p$-value in
%\eq{eq:naive-p}, 
Thus, we employ the following \emph{conditional}
$p$-value 
\begin{align}
 \label{eq:selective_p}
 p^{\rm selective}_k
  = 
 \PP_{{\rm H}_{0,k}}
 \Big(
 |\bm \eta_k^\top \bm X| \ge |\bm \eta_k^\top \bm x^{\rm obs}|
 ~\large|~
 \cA(\bm X) = \cA(\bm x^{\rm obs}), 
 \bm q(\bm X) = \bm q(\bm x^{\rm obs})
 \Big),
\end{align}
where
$ \cA(\bm X) = \cA(\bm x^{\rm obs})$
indicates
the event that the detected CP vector for a random sequence $\bm X$ is the same as the detected CP vector for the observed sequence $\bm x^{\rm obs}$. 
The second condition 
$\bm q(\bm X)=\bm q(\bm x^{\rm obs})$
indicates that
the component which is independent of the test statistic $\bm \eta_k^\top \bm X$ for a random sequence $\bm X$ is the same as the one for 
%the observed sequence 
$\bm x^{\rm obs}$
\footnote{
In the unconditional case \eq{eq:naive-p},
 the condition
 $\bm q(\bm X)=\bm q(\bm x^{\rm obs})$
 does not change the sampling distribution 
 since
 $\bm \eta_k^\top \bm X$
 and
 $\bm q(\bm X)$
 are (marginally) independent.
 On the other hand,
 under the condition with $\cA(\bm X)=\cA(\bm x^{\rm obs})$,
 $\bm \eta_k^\top \bm X$
 and
 $\bm q(\bm X)$
 are not conditionally independent.
 %
% Thus, it is necessary to
% additionally condition on 
% $\bm q(\bm X)=\bm q(\bm x^{\rm obs})$.
 %
 See \citet{fithian2014optimal, lee2016exact} for the details. 
 }.
The ${\bm q}(\bm X)$ corresponds to the component $\bm z$ in the seminal paper (see \cite{lee2016exact}, Sec 5, Eq 5.2 and Theorem 5.2), and it is given by
\begin{align*}
 \bm q(\bm X) = (I_N - \bm c \bm \eta_k^\top)\bm X
 ~
 \text{ where }
 \bm c = \Sigma \bm \eta_k(\bm \eta_k^\top \Sigma \bm \eta_k)^{-1}.
\end{align*}
The $p$-value in \eq{eq:selective_p} is called 
\emph{selective type I error} or 
\emph{selective $p$-values} in SI literature~\cite{fithian2014optimal}.
Figures~\ref{fig:naive_p_value_distribution} and \ref{fig:selective_p_value_distribution}
in
Appendix~\ref{appendix:distribution_p_value}
show the distribution of
naive $p$-values
and
selective $p$-values
when the null hypothesis ${\rm H}_{0,k}$
is true.
The naive $p$-values are not uniformly distributed, while selective $p$-values are. 
The uniformly distributed property is necessary for valid $p$-values since it indicates
\begin{align*}
 \PP_{\rm H_{0,k}}
 \left(
 p^{\rm selective}_k < \alpha 
 \right)
 = \alpha,
 ~~~
 \forall
 \alpha \in [0, 1].
\end{align*}
Our contribution is to provide an efficient method for computing selective $p$-value in \eq{eq:selective_p} by characterizing the selection event $\cA(\bm X)=\cA(\bm x^{\rm obs})$, which is computationally challenging because we have to find the whole set of sequences in $\RR^N$ having the same optimal CP vectors on $\bm x^{\rm obs}$.
% as we will see in the following sections. 

\section{Proposed Method} \label{sec:sec3}

%\begin{figure}[t]
%   \begin{minipage}{0.635\textwidth}
%     \centering
%     \includegraphics[width=0.78\linewidth]{schematic_illustration}
%     \caption{
%Schematic illustration of the proposed OptSeg-SI method. 
%%
%By applying a CP detection algorithm on the observed sequence $\bm x^{\rm obs}$, the optimal CP vector $\bm \tau^{\rm det}$ is obtained.
%%
%%
%In the OptSeg-SI method, the statistical inference is conducted conditional on the subspace $\cX$ whose data has the same optimal CP vector as $\bm x^{\rm obs}$.
%%
%We introduce a parametric programing method for efficiently characterizing the conditional data space $\cX$.
%}
%   \label{fig:schematic_illustration}
%   \end{minipage}\hfill
%   \begin{minipage}{0.345\textwidth}
%     \centering
%     \includegraphics[width=\linewidth]{piecewise}
%     \caption{A set of QFs each of which corresponds to a CP vector
%$\bm \tau \in \cT_{k, n}$. The dotted grey QFs correspond to CP vectors that are not optimal for any $z \in \RR$. A set $\{\bm t_1, \bm t_2, \bm t_3, \bm t_4\}$ contains CP vectors that are \emph{optimal} for some $z \in \RR$.
%}
%   \label{fig:piecewise}
%   \end{minipage}
%\end{figure}

\begin{figure}[t]
   \begin{minipage}{0.635\textwidth}
     \centering
     \includegraphics[width=0.78\linewidth]{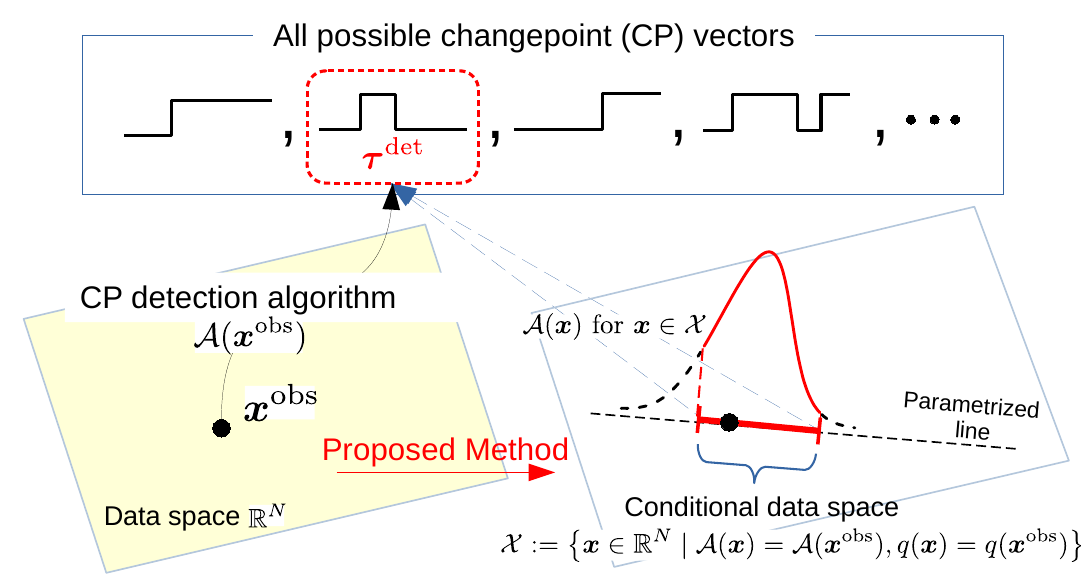}
     \caption{
Schematic illustration of the proposed OptSeg-SI method. 
By applying a CP detection algorithm on the observed sequence $\bm x^{\rm obs}$, the optimal CP vector $\bm \tau^{\rm det}$ is obtained.
In the OptSeg-SI method, the statistical inference is conducted conditional on the subspace $\cX$ whose data has the same optimal CP vector as $\bm x^{\rm obs}$.
We introduce a parametric programming method for efficiently characterizing the conditional data space $\cX$.
}
   \label{fig:schematic_illustration}
   \end{minipage}\hfill
   \begin{minipage}{0.345\textwidth}
     \centering
     \includegraphics[width=\linewidth]{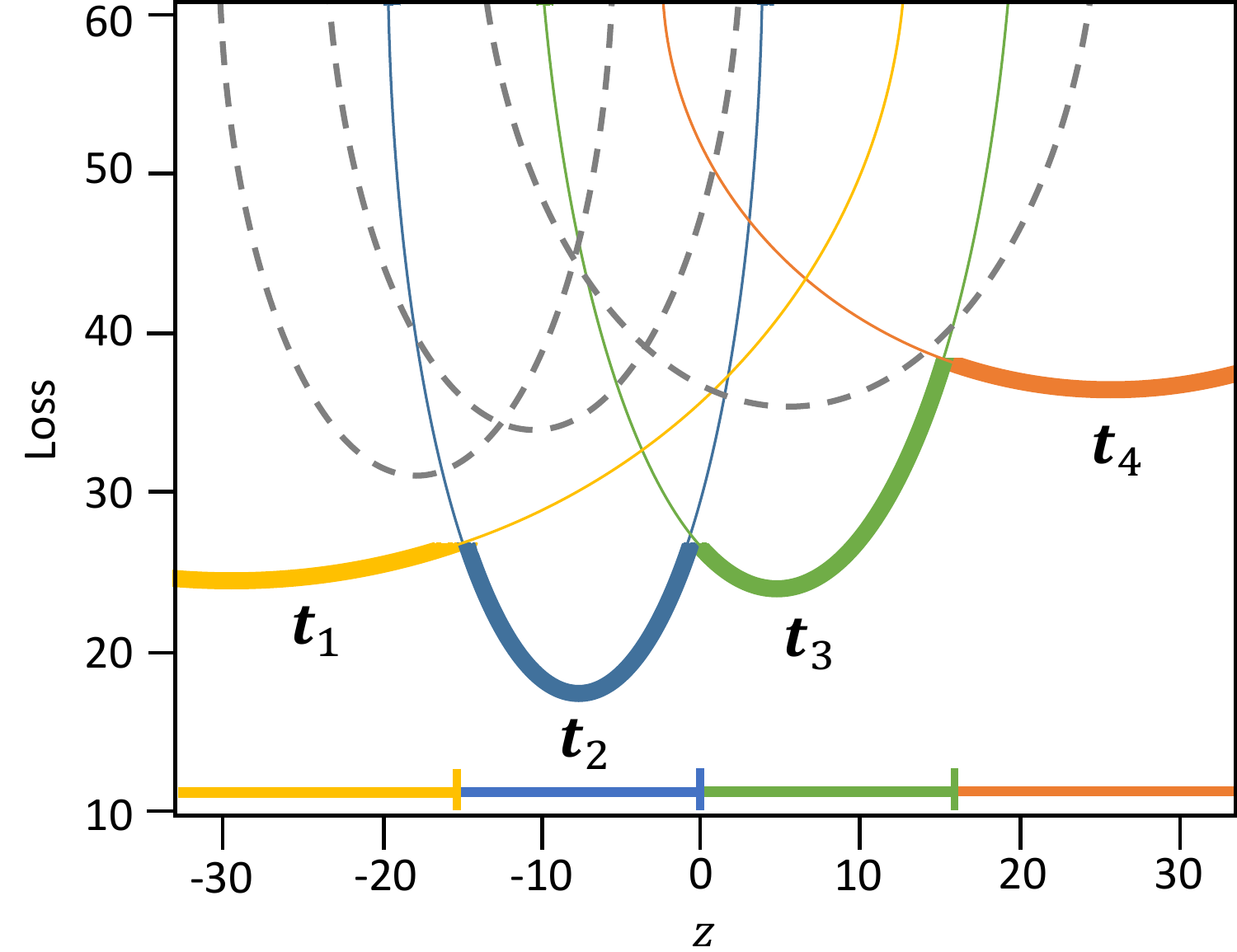}
     \caption{A set of QFs each of which corresponds to a CP vector
$\bm \tau \in \cT_{k, n}$. The dotted grey QFs correspond to CP vectors that are not optimal for any $z \in \RR$. A set $\{\bm t_1, \bm t_2, \bm t_3, \bm t_4\}$ contains CP vectors that are \emph{optimal} for some $z \in \RR$.
}
   \label{fig:piecewise}
   \end{minipage}
\end{figure}

%
We propose a method for computing selective $p$-values in \eq{eq:selective_p}.
%for detected optimal CPs.
%
We focus here on the case where the number of CPs $K$ is fixed.
The case for unknown $K$ will be discussed in \S \ref{sec:sec4}.
Figure \ref{fig:schematic_illustration} shows the schematic illustration of the OptSeg-SI method. 

\subsection{Conditional Data Space Characterization}
Let us define the set of 
%sequences
$\bm x \in \RR^N$ which satisfies the conditions %of the selective $p$-value 
in 
\eq{eq:selective_p} by 
\begin{align*}
 \cX = \{\bm x \in \RR^N ~\mid~ \cA(\bm x)=\cA(\bm x^{\rm obs}), \bm q(\bm x)=\bm q(\bm x^{\rm obs})\}. 
\end{align*}
Based on the second condition 
$\bm q(\bm x)=\bm q(\bm x^{\rm obs})$, the data in $\cX$ is restricted to a line (see Sec 6 in \cite{liu2018more}, and \cite{fithian2014optimal}).
Therefore, the set $\cX$ can be re-written,
using a scalar parameter $z \in \RR$, as
\begin{align*} 
 \cX = \{\bm a + \bm b z \mid z \in \cZ\}, \text{ where }
 \cZ = \{z \in \RR \mid \cA(\bm a + \bm b z)=\cA(\bm x^{\rm obs})\}
\end{align*}
with
$
\bm a = \bm q(\bm x^{\rm obs})
$
and
$
\bm b = \Sigma \bm \eta_k(\bm \eta_k^\top \Sigma \bm \eta_k)^{-1}
$.
Now,
let us denote a random variable $Z \in \RR$ and its observation $z^{\rm obs} \in \RR$,
which satisfy 
$
\bm X = \bm a + \bm b Z
\text{ and }
\bm x^{\rm obs} = \bm a + \bm b z^{\rm obs}
$.
%for the random sequence
%$\bm X$
%and
%its observation
%$\bm x^{\rm obs}$. 
%
Then, 
the selective $p$-value in \eq{eq:selective_p} 
is re-written as
\begin{align}
% \label{eq:selective_p_1d}
\begin{aligned}
 p^{\rm selective}_k
 =
 \PP_{{\rm H}_{0,k}}
 \left(
 |\bm \eta_k^\top \bm X|
 >
 |\bm \eta_k^\top \bm x^{\rm obs}|
 \mid
 \bm X \in \cX
 \right)
 \label{eq:selective_p_1d2}
 =
 \PP_{{\rm H}_{0,k}}
 \left(
 |Z| > |z^{\rm obs}| \mid Z \in \cZ
 \right).
\end{aligned}
\end{align}
Since variable $Z \sim \NN(0, \bm \eta_k^\top \Sigma \bm \eta_k)$ under the null hypothesis,
%the conditional distribution
%$\PP_{{\rm H}_{0,k}} \left(
%Z \mid Z \in \cZ
%\right)$
%follows a truncated Normal distribution.
%with the truncation region
%$\cZ$. 
%
the law of 
$Z \mid Z \in \cZ$
follows a truncated Normal distribution.
Once the truncation region
$\cZ$ 
is identified,
the selective $p$-value in \eq{eq:selective_p_1d2} can be computed as
%\begin{align}
% \label{eq:selective_p_final}
% \hspace{-2.33mm}p^{\rm selective}_k
% =
% 2
% \min
% \Big\{
% F^{\cZ}_{0, \bm \eta_k^\top \Sigma \bm \eta_k}(z),
% 1 -  F^{\cZ}_{0, \bm \eta_k^\top \Sigma \bm \eta_k}(z)
% \Big\},
%\end{align}
\begin{align*}
 \label{eq:selective_p_final}
% \hspace{-2.33mm}
 p^{\rm selective}_k
 =
 F^{\cZ}_{0, \bm \eta_k^\top \Sigma \bm \eta_k}(-|z^{\rm obs}|) + 1 -  F^{\cZ}_{0, \bm \eta_k^\top \Sigma \bm \eta_k}(|z^{\rm obs}|),
\end{align*}
where
$
F^{\cE}_{m, s^2}
$
is the c.d.f. of the truncated Normal distribution with
mean
$m$,
variance
$s^2$
and
the truncation region
$\cE$. 
Therefore, the main task is to identify $\cZ$.

\paragraph{Important notations.}
In the rest of this paper,
we use the following notations. 
Since we focus on a set of sequences
parametrized by a scalar parameter
$z \in \RR$,
we denote these sequences by 
\begin{equation} \label{eq:x_z}
\bm x(z) = \bm a + \bm b z
\end{equation}
or
just simply by 
$z$.
For a sequence with length $n \in [N]$, 
the set of all possible CP vectors with dimension $k \in [K]$ is written as 
$\cT_{k, n}$.
Given
$\bm x(z)$,
the loss of segmenting its first $n$ sub-sequence 
$\bm x(z)_{1:n}$ 
with a $k$-dimensional CP vector
$\bm \tau \in \cT_{k, n}$
is written as
$
% \label{eq:loss}
 L_{k, n}(z, \bm \tau)
 =
 \sum_{\kappa=1}^{k+1}
 C\left(
 \bm x(z)_{\tau_{\kappa -1}+1:\tau_{\kappa}}
 \right).
$
For a subsequence 
$\bm x(z)_{1:n}$,
the optimal loss and 
the optimal $k$-dimensional CP vector are respectively written as
\begin{align}
 \label{eq:loss_min_and_cp_min}
 L^{\rm opt}_{k, n}(z) = \min_{\bm \tau \in \cT_{k, n}} L_{k, n}(z, \bm \tau), \quad
 \bm T^{\rm opt}_{k, n}(z) = \arg \min_{\bm \tau \in \cT_{k, n}} L_{k, n}(z, \bm \tau).
\end{align}
Note that
the notation
$z \in \RR$
in the definition \eq{eq:loss_min_and_cp_min}
indicates that
it corresponds to the sequence
$\bm x(z)$.
%$ = \bm a + \bm b z$.

\paragraph{Main idea for identifying truncation region $\cZ$.}
Since we denoted $\bm x(z) = \bm a + \bm b z$ as in \eq{eq:x_z}, truncation region $\cZ$ is re-written as follows
\begin{align}
\label{eq:cZ1}
\begin{aligned}
	\cZ = \{z \in \RR \mid \cA(\bm x(z))=\cA(\bm x^{\rm obs})\}
		= \{z \in \RR \mid \bm T^{\rm opt}_{K,N}(z)=\cA(\bm x^{\rm obs})\}.
\end{aligned}
\end{align}
The main idea 
%behind the proposed method 
is to efficiently compute the optimal path of CP vectors 
$\bm T^{\rm opt}_{K, N}(z) \in \cT_{K, N}$
for all 
values of $z \in \RR$, which is computationally challenging. 
After $\bm T^{\rm opt}_{K, N}(z)$ is identified for all $z \in \RR$, truncation region $\cZ$ can be easily characterized, and the selective $p$-value in \eq{eq:selective_p_1d2} can be computed. 

\subsection{Parametric CP detection}
We introduce an efficient way to compute $\bm T^{\rm opt}_{K, N}(z)$ for all $z \in \RR$.
Although it seems intractable to solve 
this problem
for infinitely many values of $z$, we can complete the task with a finite number of operations. 
Algorithm~\ref{alg:paraCP} shows the overview of
our parametric CP detection method.
Here,
the algorithm is described in terms of general
$n \in [N]$
and
$k \in [K]$
along with
a set of CP vectors
$\hat{\cT}_{k, n}$. 
In the current subsection,
we
set 
$n=N$,
$k=K$
and
$\hat{\cT}_{k, n}=\cT_{k, n}$. 
The case
with
general 
$n$,
$k$
and
$\hat{\cT}_{k, n}$
will be discussed in \S3.3.

In our parametric CP detection method, 
we exploit the fact that, 
for each CP vector $\bm \tau \in \cT_{k, n}$,
the loss function is written as a quadratic function (QF) of $z$ 
whose
coefficients 
depend on
$\bm \tau \in \cT_{k, n}$.
Since the number of possible CP vectors in
$\cT_{k, n}$
is finite,
the parametric CP detection problem can be characterized
by a finite number of these QFs.
Figure~\ref{fig:piecewise}
illustrates
the set of QFs
each of which
corresponds to
a CP vector
$\bm \tau \in \cT_{k, n}$.
Since the  
minimum loss
for each $z \in \RR$ is the point-wise minimum of these QFs, 
the optimal loss function 
$L^{\rm opt}_{k, n}(z)$ 
is the lower envelope of the set of QFs, 
which is represented as a piecewise QF of $z \in \RR$.
Parametric CP detection is interpreted as the problem of identifying this piecewise QF. 

In Algorithm~\ref{alg:paraCP},
multiple 
\emph{breakpoints}
$
 z_1 < z_2 < \ldots < z_U 
$
%\begin{center}
%\vspace{-7pt}
%$ z_1 < z_2 < \ldots < z_U $
%\vspace{-7pt}
%\end{center}
%on a line
%$
%z \in \RR
%$
are computed one by one.
Each breakpoint 
$z_u, u \in [U],$
indicates a point at which
the optimal CP vector
is replaced from one to the other in the piecewise QF. 
By finding all these breakpoints
$\{z_u\}_{u=1}^U$
and
the optimal CP vectors
$\{\bm t_u\}_{u=1}^U$,
%between all the two adjacent breakpoints, 
the piecewise QF
as %depicted 
in Figure~\ref{fig:piecewise} 
can be identified. 

The algorithm is initialized at the optimal CP vector
for 
$z = -\infty$,
which can be easily identified based on the coefficients of the QFs. 
At step
$u, u \in [U],$
the task is to find the next breakpoint
$z_{u+1}$
and the next optimal CP vector
$\bm t_{u+1}$.
This task can be 
%simply 
done 
by finding the smallest
$z_{u+1}$
greater than
$z_u$
among the intersections of the current QF
$L_{k, n}(z, \bm t_u)$
and
each of the other QFs
$L_{k, n}(z, \bm \tau)$
for
$\bm \tau \in \cT_{k, n} \setminus \{\bm t_u\}$. 
This step is repeated until we find the optimal CP vector when $z = +\infty$.
%which can be identified based on the coefficients of the QFs.
%
The algorithm returns the sequences of breakpoints and optimal CP vectors $\{(z_u, \bm t_u)\}_{u=1}^U$. 
The entire path of optimal CP vectors for $z \in \RR$ 
is given by 
$
	\bm T^{\rm opt}_{k, n}(z) = \bm t_u, u \in [U], \text{ if } z \in [z_u, z_{u+1}].
$
%\begin{align*}
% \bm T^{\rm opt}_{k, n}(z)
% =
% \mycase{
% \bm t_1 & \text{ if } z \in (z_1 = -\infty, z_2], \\
% \bm t_2 & \text{ if } z \in [z_2, z_3], \\
% ~~~~~ \vdots & \\
% \bm t_U & \text{ if } z \in [z_U, z_{U+1} = +\infty).
% }
%\end{align*}

\begin{algorithm}[t]
\renewcommand{\algorithmicrequire}{\textbf{Input:}}
\renewcommand{\algorithmicensure}{\textbf{Output:}}
\begin{footnotesize}
 \begin{algorithmic}[1]
  \REQUIRE $n$, $k$, $\hat{\cT}_{k, n}$
%  \vspace{2pt}
  \STATE $u \lA 1$, $z_1 \lA -\infty$, 
  $\bm t_1 \lA \bm T^{\rm opt}_{k, n}(z_u) = \arg \min \limits_{\bm \tau \in \hat{\cT}_{k, n}} L_{k, n}(z_u, \bm \tau)$
  \WHILE { $z_u < +\infty$}
%  \vspace{2pt}
  \STATE Find the next breakpoint $z_{u+1} > z_u$ and the next optimal CP vector $\bm t_{u+1}$ such that 
%  \vspace{2pt}
  \begin{center}
   $L_{k, n}(z_{u+1}, \bm t_u) = L_{k, n}(z_{u+1}, \bm t_{u+1}).$
  \end{center}
%  \vspace{2pt}
  \STATE $u \lA u+1$
%  \vspace{2pt}
  \ENDWHILE
%  \vspace{2pt}
  \STATE $U \lA u$
%  \vspace{2pt}
  \ENSURE $\{(z_u, \bm t_u)\}_{u=1}^U$
 \end{algorithmic}
\end{footnotesize}
\caption{{\tt paraCP}($n, k, \hat{\cT}_{k, n}$)}
\label{alg:paraCP}
\end{algorithm}

\subsection{Parametric DP}
Unfortunately,
parametric CP detection algorithm
with the inputs
$N$, $K$ and 
$\cT_{K, N}$ 
in the previous subsection
is computationally impractical  
because the number of all possible CP vectors
$|\cT_{K, N}|$
is exponentially increasing with
$N$ and $K$. 
To resolve this computational issue,
we 
utilize the concept of standard DP,
and apply
to parametric case,
which
we call
\emph{parametric DP}.
The basic idea of parametric DP is to 
exclude
the CP vectors
$\bm \tau \in \cT_{k, n}$
that cannot be optimal at any
$z \in \RR$.
%by introducing Bellman equation for parametric case. 

\paragraph{Standard DP (specific value of $z$).}
In standard DP for a CP detection problem (for a specific $z$) with $N$ and $K$,
we use $K \times N$ table 
whose
$(k,n)^{\rm th}$
element contains
$\bm T^{\rm opt}_{k, n}(z)$,
the vector of optimal $k$ CPs for the subsequence
$\bm x(z)_{1:n}$. 
The optimal CP vector
for each of the subproblem with $n$ and $k$
% in the table 
can be used for efficiently computing the optimal CP vector for the original problem with $N$ and $K$. 

Let
${\tt concat}(\bm v, s)$
be the operator for concatenating
a vector
$\bm v$
and
a scalar
$s$. 
Then, 
%in standard DP (for a single $z$),
it is known that the following equation,
which is often called
\emph{Bellman equation}, 
holds: 
\begin{align}
 \label{eq:Bellman1}
 \bm T^{\rm opt}_{k, n}(z)
 = 
 \arg \min_{\bm \tau(m)}
 \left\{
 L_{k, n}
 \left(
 z, \bm \tau(m)
 \right)
 \right\}_{m=k}^{n-1}, 
\end{align}
where
$
 \bm \tau(m) = {\tt concat}(\bm T^{\rm opt}_{k-1, m}(z), m), m \in \{k, \ldots, n-1\}. 
$
The Bellman equation
\eq{eq:Bellman1}
enables us to efficiently compute the optimal
CP vector for the problem with 
$n$ and $k$
by using the optimal CP vectors of its sub-problems.
%for the problems with 
%$m \in \{k, \ldots, n\}$ 
%and 
%$k-1$.

\paragraph{Parametric DP (for all values of $z \in \RR$).}
Our basic idea is to similarly construct a
$K \times N$
table
whose
$(k, n)^{\rm th}$
element contains
$
 \cT^{\rm opt}_{k, n}
 =
 \left\{
 \bm \tau \in \cT_{k, n}
 \mid
 {^\exists}z \in \RR
 \text{ s.t. }
 L^{\rm opt}_{k, n}(z) = L_{k, n}(z, \bm \tau) 
\right\}
$,
which is a \emph{set of CP vectors} that are optimal for some $z \in \RR$. 
To identify $\cT^{\rm opt}_{k, n}$, we construct a set $\hat{\cT}_{k, n} \supseteq \cT^{\rm opt}_{k, n}$, which is a set of CP vectors having potential to be optimal.
%
%Then, it is obvious that $\cT^{\rm opt}_{k, n} \subseteq \hat{\cT}_{k, n}$.
%
In the same way as
\eq{eq:Bellman1},
we can consider Bellman equation for constructing $\hat{\cT}_{k, n}$
as described in the following Lemma.

%\begin{algorithm}[t]
%\renewcommand{\algorithmicrequire}{\textbf{Input:}}
%\renewcommand{\algorithmicensure}{\textbf{Output:}}
%\begin{footnotesize}
% \begin{algorithmic}[1]
%  \REQUIRE $\bm x(z)$ and $K$
%  \vspace{2pt}
%  \FOR{$k=1$ to $K$}
%  \vspace{2pt}
%  \FOR{$n=1$ to $N$}
%  \vspace{2pt}
%  \STATE $\hat{\cT}_{k,n}$ $\lA$ Eq.\eq{eq:T_Bellman}
%  \vspace{2pt}
%  \STATE $\{(z_u, \bm t_u)\}_{u=1}^U$ $\lA$ {\tt paraCP}($n, k, \hat{\cT}_{k, n}$)
%  \vspace{2pt}
%  \STATE $\cT^{\rm opt}_{k,n} \lA \{\bm t_u\}_{u=1}^U$ 
%  \vspace{2pt}
%  \ENDFOR
%  \vspace{2pt}
%  \ENDFOR
%  \vspace{2pt}
%  \ENSURE $\cT^{\rm opt}_{K,N}$
% \end{algorithmic}
%\end{footnotesize}
%\caption{{\tt paraDP}($\bm x(z), K$)}
%\label{alg:para_DP}
%\end{algorithm}

%=========== ParaDP ===========

\begin{lemma} \label{lemma:bellman_fixed_k}
For $n \in [N]$ and $k \in [K]$,
the set of CP vectors 
having potential to be optimal is
constructed as
%\begin{align}
% \label{eq:T_Bellman}
$
 \hat{\cT}_{k, n} = {\mathop{\cup}}_{m=k}^{n-1} \{{\tt concat}({\cT}^{\rm opt}_{k-1, m}, m)\},
$
%\end{align}
where
we extend the 
${\tt concat}$
operator
for the case where the first argument is a set of vectors,
which simply returns the set of concatenated vectors. 
\end{lemma}
In other words, the set $\hat{\cT}_{k, n} $ can be generated from the optimal CP vectors of its sub-problems ${\cT}^{\rm opt}_{k-1, m}$ 
for
$m \in \{k, \ldots, n-1\}$.
The proof for this result is deferred to Appendix \ref{appendix:proof_lemma_1}.
%
%\noindent{\bf Proof}.
%\begin{proof}
%We prove the lemma by showing that
%any CP vector
%$\bm \tau \not \in {\cT}^{\rm opt}_{k-1, m}$,
%for
%$m \in \{k, \ldots, n-1\}$, 
%cannot be subvector of the optimal CP vectors for problems with larger 
%$n$ and $k$ 
%for 
%any
%$z \in \RR$,
%i.e., ${\tt concat}(\bm \tau, m) \not \in {\cT}^{\rm opt}_{k, n} $ for $n > m$.
%%
%For
%$m \in \{k, \ldots, n-1\}$,
%let
%$\bm \tau \not \in {\cT}^{\rm opt}_{k-1, m}$
%be a CP vector
%which is NOT optimal for all 
%$z \in \RR$,
%i.e.,
%\begin{align*}
% L_{k-1, m}(z, \bm \tau) > L^{\rm opt}_{k-1, m}(z)
% ~~~
% \forall z \in \RR.
%\end{align*}
%%
%It suggests that,
%for any
%$m \in \{k, \ldots, n-1\}$
%and
%$z \in \RR$, 
%\begin{align*}
% L^{\rm opt}_{k, n}(z) 
% &=
% \min_{m^\prime \in \{k, \ldots, n-1\}}
% \left(
% L^{\rm opt}_{k-1, m^\prime}(z)
% +
% C(\bm x(z)_{m^\prime + 1:n})
% \right)
% \\
% &
% \le
% L^{\rm opt}_{k-1, m}(z) +  C(\bm x(z)_{m+1:n})
% \\
% &
% <
% L_{k-1, m}(z, \bm \tau) + C(\bm x(z)_{m+1:n})
%\end{align*}
%for all $z \in \RR$. 
%%
%Thus, 
%for any choice of $m \in \{k, \ldots, n-1\}$ and $z \in \RR$, 
%$\bm \tau \not \in {\cT}^{\rm opt}_{k-1, m}$
%cannot be a subvector of the optimal CP vector
%for problems with larger $n$ and $k$.
%%
%In other words, 
%only the CP vectors in
%$\cup_{m=k}^{n-1} {\cT}^{\rm opt}_{k-1, m}$
%can be used as the subvector of optimal CP vectors for problems with larger $n$ and $k$. 
%\end{proof}
%
%
From Lemma \ref{lemma:bellman_fixed_k}, we can efficiently construct $\hat{\cT}_{k, n}$
which is subsequently used to identify $\cT^{\rm opt}_{k, n}$.
By repeating the recursive procedure and storing 
$\cT^{\rm opt}_{k, n}$
in the $(k, n)^{\rm th}$ element of the table 
from smaller $n$ and $k$ to larger $n$ and $k$, 
we can end up with
$\hat{\cT}_{K, N} \supseteq \cT^{\rm opt}_{K, N}$.
%which surely contains all the CP vectors that could be optimal
%for some $z \in \RR$.
%
By using parametric DP,
the size of
$\hat{\cT}_{K, N}$
can be smaller than the size of all possible CP vectors
$\cT_{K, N}$,
which makes
the computational cost of
${\tt paraCP}(N, K, \hat{\cT}_{K, N})$ substantially decreased
compared to
${\tt paraCP}(N, K, \cT_{K, N})$.

The parametric DP method is presented in Algorithm~\ref{alg:para_DP}
and the entire OptSeg-SI method for computing selective $p$-values of the optimal CPs is summarized in Algorithm~\ref{alg:si_opt_cp}.
Although they are not explicitly described in the algorithm, we also used several computational tricks for further reducing the size of $\hat{\cT}_{k, n}$. See Appendix \ref{appendix:computational_trick} for the details.

%=========== ParaDP ===========

\begin{figure}[t]
\begin{minipage}{0.45\textwidth}
\centering
\begin{algorithm}[H]
\renewcommand{\algorithmicrequire}{\textbf{Input:}}
\renewcommand{\algorithmicensure}{\textbf{Output:}}
\begin{footnotesize}
 \begin{algorithmic}[1]
  \REQUIRE $\bm x(z)$ and $K$
  \vspace{2pt}
  \FOR{$k=1$ to $K$}
  \vspace{2pt}
  \FOR{$n=1$ to $N$}
  \vspace{2pt}
  \STATE $\hat{\cT}_{k,n}$ $\lA$ Lemma \ref{lemma:bellman_fixed_k}
  \vspace{1pt}
  \STATE $\{(z_u, \bm t_u)\}_{u=1}^U$ $\lA$ {\tt paraCP}($n, k, \hat{\cT}_{k, n}$)
  \vspace{1pt}
  \STATE $\cT^{\rm opt}_{k,n} \lA \{\bm t_u\}_{u=1}^U$ 
  \vspace{2pt}
  \ENDFOR
  \vspace{1pt}
  \ENDFOR
  \vspace{1pt}
  \ENSURE $\cT^{\rm opt}_{K,N}$
 \end{algorithmic}
\end{footnotesize}
\caption{{\tt paraDP}($\bm x(z), K$)}
\label{alg:para_DP}
\end{algorithm}
\end{minipage}\hfill
\begin{minipage}{0.52\textwidth}
\centering
\begin{algorithm}[H]
\renewcommand{\algorithmicrequire}{\textbf{Input:}}
\renewcommand{\algorithmicensure}{\textbf{Output:}}
\begin{footnotesize}
 \begin{algorithmic}[1]
  \REQUIRE $\bm x_{\rm obs}$ and $K$
  \vspace{1pt}
  \STATE $\bm \tau^{\rm det} \lA \cA(\bm x^{\rm obs})$
  \vspace{1pt}
  \FOR{$\tau^{\rm det}_k \in \bm \tau^{\rm det}$}
  \vspace{1pt}
  \STATE $\bm x(z)$ $\lA$ Eq.\eq{eq:x_z} 
  \vspace{1pt}
  \STATE $\cT^{\rm opt}_{K,N}$ $\lA$ {\tt paraDP}($\bm x(z)$, $K$)
%  \vspace{2pt}
  \STATE $\cZ \lA \mathop{\cup}_{\bm T^{\rm opt}_{K,N}(z) \in \cT^{\rm opt}_{K,N}} \{z: \bm T^{\rm opt}_{K,N}(z) = \cA(\bm x^{\rm obs})\}$
%  \vspace{2pt}
  \STATE $p^{\rm selective}_k$ $\lA$ Eq.\eq{eq:selective_p_1d2}
%  \vspace{2pt}
  \ENDFOR
%  \vspace{2pt}
  \ENSURE $\{(\tau^{\rm det}_k, p^{\rm selective}_k)\}_{k=1}^K$
 \end{algorithmic}
\end{footnotesize}
\caption{SI for Optimal CPs (OptSeg-SI)}
\label{alg:si_opt_cp}
\end{algorithm}

\end{minipage}

\end{figure}

%\begin{algorithm}[t]
%\renewcommand{\algorithmicrequire}{\textbf{Input:}}
%\renewcommand{\algorithmicensure}{\textbf{Output:}}
%\begin{footnotesize}
% \begin{algorithmic}[1]
%  \REQUIRE $\bm x_{\rm obs}$ and $K$
%  \vspace{2pt}
%  \STATE $\bm \tau^{\rm det} \lA \cA(\bm x^{\rm obs})$
%  \vspace{2pt}
%  \FOR{$\tau^{\rm det}_k \in \bm \tau^{\rm det}$}
%  \vspace{2pt}
%  \STATE $\bm x(z)$ $\lA$ Eq.\eq{eq:x_z} 
%  \vspace{2pt}
%  \STATE $\cT^{\rm opt}_{K,N}$ $\lA$ {\tt paraDP}($\bm x(z)$, $K$)
%  \vspace{2pt}
%  \STATE $\cZ \lA \mathop{\cup}_{\bm T^{\rm opt}_{K,N}(z) \in \cT^{\rm opt}_{K,N}} \{z: \bm T^{\rm opt}_{K,N}(z) = \cA(\bm x^{\rm obs})\}$
%  \vspace{2pt}
%  \STATE $p^{\rm selective}_k$ $\lA$ Eq.\eq{eq:selective_p_1d2}
%  \vspace{2pt}
%  \ENDFOR
%  \vspace{2pt}
%  \ENSURE $\{(\tau^{\rm det}_k, p^{\rm selective}_k)\}_{k=1}^K$
% \end{algorithmic}
%\end{footnotesize}
%\caption{SI for Optimal CPs (OptSeg-SI)}
%\label{alg:si_opt_cp}
%\end{algorithm}

\section{Extension to Unknown $K$ Case} \label{sec:sec4}
We present an approach for testing the significance of CPs detected by 
\eq{eq:opt2}.
The basic idea is the same as the proposed method for fixed $K$.
With a slight abuse of notations, we use the following similar notations as the fixed $K$ case.
For a sequence with length $n \in [N]$, 
the set of all possible CP vectors is written as 
$\cT_{n}$.
Given $\bm x(z)$ as in \eq{eq:x_z},
the loss of segmenting its sub-sequence 
$\bm x(z)_{1:n}$ 
with a CP vector 
$\bm \tau \in \cT_{n}$
is written as
$
 L_{n}(z, \bm \tau)
 =
 \sum_{\kappa=1}^{{\rm dim}(\bm \tau) + 1}
 C\left(
 \bm x(z)_{\tau_{\kappa-1}+1:\tau_{\kappa}}
 \right) + \beta {\rm dim}(\bm \tau).
$
The optimal loss and 
the optimal CP vector on $\bm x(z)_{1:n}$  are respectively written as
$
% \label{eq:unknon_k_loss_min}
 L^{\rm opt}_{n}(z) =  \min_{\bm \tau \in \cT_{n}} L_{n}(z, \bm \tau)
$,
$
 \bm T^{\rm opt}_{n}(z) = \arg \min_{\bm \tau \in \cT_{n}} L_{n}(z, \bm \tau) \footnote{We recently noticed that $\ell_1$-penalty based SI for CP detection was extended to $\ell_0$-penalty~\cite{jewell2019testing}, which results in a similar approach with the ``unknown $K$ case'' in our algorithm.}.
$
\paragraph{Identification of truncation region $\cZ$.} 
To calculate $p^{\rm selective}_k$ for the $k^{\rm th}$ detected CP,
we characterize the truncation region
%also need to compute truncation region $\cZ$. As similar to Section \ref{sec:sec3},
%\begin{align}
% \label{eq:cZ2}
% \cZ = \{z \in \RR \mid \bm T^{\rm opt}_{N}(z) = \cA(\bm x^{\rm obs}) \},
%\end{align}
$
 \cZ = \{z \in \RR \mid \bm T^{\rm opt}_{N}(z) = \cA(\bm x^{\rm obs}) \},
$
by computing
$\bm T^{\rm opt}_N(z)$
for all
$z \in \RR$. 
%then the remaining task is to compute $\bm T^{\rm opt}_{N}(z)$ for all $z \in \RR$.
We can slightly modify Algorithm~1 to the unknown $K$ case to compute $\bm T^{\rm opt}_{N}(z)$ for all $z \in \RR$. % based on the set $\cT_N$ by following \eq{eq:unknon_k_cp_min}.
Let $\cT^{\rm opt}_n$ denote a set of CP vectors that are optimal at some $z \in \RR$ for subsequence $\bm x(x)_{1:n}$ as 
$
 \cT^{\rm opt}_{n}
 =
 \left\{
 \bm \tau \in \cT_{n}
 \mid
 {^\exists}z \in \RR
 \text{ s.t. }
 L^{\rm opt}_{n}(z) = L_{n}(z, \bm \tau) 
\right\}.
$

Since the set of all possible CP vectors
$\cT_N$
is huge,
we use parametric DP
with two additional computational tricks (Lemmas 2 and 3 below) 
for finding a substantially reduced set of CP vectors
$\hat{\cT}_N \subseteq \cT_N$
which contains all the optimal CP vectors for any $z \in \RR$,
i.e.,
$\hat{\cT}_N \supseteq T^{\rm opt}_N$. 
%
%However, the $\cT_N$ is huge, leading to intractable computation.
%
%Therefore, we use the parametric DP and additionally propose pruning lemma for substantially reducing the size of this set.
%
%Since the set $\cT_{n}$ is large, we want to construct a smaller set $\hat{\cT}_n \subset \cT_{n}$ that can be used to efficiently compute $\cT^{\rm opt}_{n}$.
%
The following two lemmas show how to construct $\hat{\cT}_n$ by removing the CP vectors that never belong to $\cT^{\rm opt}_{n}$.
\begin{lemma}\label{lemma:unknown_k_lemma_1}
{\it
For $m < n$, if a vector $\bm \tau \not \in \cT^{\rm opt}_m$, then ${\tt concat}(\bm \tau, m) \not \in \cT^{\rm opt}_n$.
}
\end{lemma}
\begin{lemma}\label{lemma:unknown_k_lemma_2}
{\it
For $m < n$,
if 
$\bm \tau \not \in \cT^{\rm opt}_m$ and 
$L_{m}(z, \bm \tau) - \beta > L^{\rm opt}_m(z) \text{ for any } z \in \RR$,
%\begin{align}
% \label{eq:eq_lemma3}
% L_{m}(z, \bm \tau) - \beta > L^{\rm opt}_m(z) \quad \forall z \in \RR, 
%\end{align}
then $\bm \tau \not \in \cT^{\rm opt}_n$.
}
\end{lemma}
\noindent
Proofs for these two lemmas are deferred to Appendix \ref{appendix:proof_lemma_2_3}. 
Based on Lemmas 2 and 3, $\hat{\cT}_n$ can be constructed by
$
 \hat{\cT}_{n} = {\mathop{\cup}}_{m=k}^{n-1} \{{\tt concat}({\cT}^{\rm opt}_{m}, m) \mathop{\cup} \cS\},
$
where $\cS$ is a set of $\bm \tau \not \in \cT^{\rm opt}_m$ that does not satisfy  Lemma 3. 
%
%We remind that the {\tt concat} operator for the case  where the first argument is a set of vectors simply returns the set of concatenated vectors between each vector in first argument and second argument. 
%
Then, we can use $\hat{\cT}_{n}$ to find $\cT^{\rm opt}_{n}$.
We store $\cT^{\rm opt}_{n}$ and continue this process recursively for larger $n$ until we get $\cT^{\rm opt}_{N}$.
After identifying $\cT^{\rm opt}_{N}$, we can fully characterize truncation region $\cZ$ and finally calculate selective $p$-values.

\section{Numerical Experiments} \label{sec:sec_exp}

%We test the performance of the OptSeg-SI method. %Here
We only highlight the main results. 
%due to space limitation. 
More details can be found in Appendix \ref{appendix:numerical_experiments}.
%
%=======================
%\subsection{Methods for Comparison} \label{subsection:method_for_comparison}

\textbf{Methods for comparison.} We compared our OptSeg-SI method with SMUCE \cite{frick2014multiscale}, which is an asymptotic test for multiple detected CPs, and SI for Binary Segmentation~\cite{hyun2018post} (BinSeg-SI).
It was reported that SI for Fused Lasso (proposed by the same authors), is worse than BinSeg-SI. Therefore, we only compared to BinSeg-SI.
We additionally compared our method with SI method for optimal CPs with  over-conditioning (OptSeg-SI-oc) to demonstrate the advantage of minimum conditioning.
The details of OptSeg-SI-oc are shown in Appendix \ref{appendix:derive_dp_selection_event} \footnote{We first developed OptSeg-SI-oc as our first SI method for optimal CPs detected by DP (unpublished). Later, its drawback (the over-conditioning) was removed by the OptSeg-SI method in this paper.
}.

%=======================

\textbf{Simulation setup.} Regarding false positive rate (FPR) experiments, we generated 1,000 null sequences $\bmx = (x_1, ..., x_N)$ in which $x_{i \in [N]} \sim \mathbb{N}(0, 1)$ for each $N \in \{10, 20, 30, 40\}$.
In regard of testing the power, we generated sequences  $\bmx = (x_1, ..., x_N)$ with sample size $N=60$, in which 
\begin{equation*}
	x_{i \in [N]} \sim \mathbb{N}(\mu_i, 1), \quad 
	\mu_i = \begin{cases}
		1 \quad \quad \quad \quad \text{ if } 1\leq i \leq 20, \\
		1 + \Delta_{\mu}  \hspace{5mm} \text{ if } 21\leq i \leq 40\\
		1 + 2 \Delta_{\mu} \quad \text{ otherwise },
	\end{cases}
\end{equation*}
for each $\Delta_{\mu} \in \{1, 2, 3, 4\}$. For each case, we ran 250
trials.
Since the tests are performed only when a CP is selected, the power is defined as follows~\cite{hyun2018post}:
\begin{equation*}
	{\rm Power\  (or \ Conditional \ Power)}  = \frac{{\rm \#\ correctly\ detected\ \&\ rejected}}{{\rm \#\ correctly\ detected}}.
\end{equation*}
A detection is considered to be correct if it is within $\pm 2$ of the true
CP locations. 
Since it is often difficult to accurately identify exact CPs in the presence of noise, many existing CP detection studies consider a detection to be correct if it is within $L$ positions of the true CP locations \cite{truong2019selective}.
We considered $L=2$ to be consistent with our competitive method \cite{hyun2018post}.
We used BIC \cite{schwarz1978estimating} for the choice of $\beta$ when $K$ is unknown.
We chose the significance level $\alpha = 0.05$. 
We used Bonferroni correction to account for the multiplicity in all the experiments.
%
%We executed the code on Intel(R) Xeon(R) CPU E5-2687W v4 @ 3.00GHz.

\begin{figure}[!t]
\begin{minipage}{0.49\textwidth}
\begin{subfigure}{.495\linewidth}
  \centering
  \includegraphics[width=\linewidth]{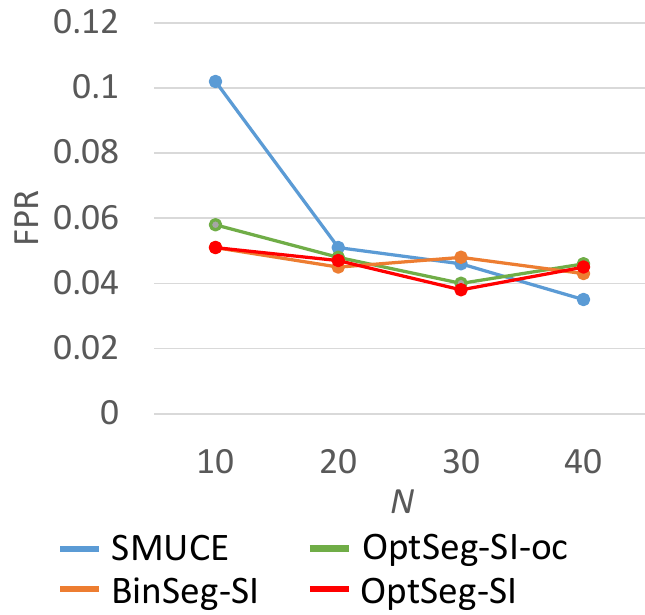}  
  \caption{False Positive Rate}
  \label{fig:fixed_k_sub_first}
\end{subfigure}
\begin{subfigure}{.495\linewidth}
  \centering
  \includegraphics[width=\linewidth]{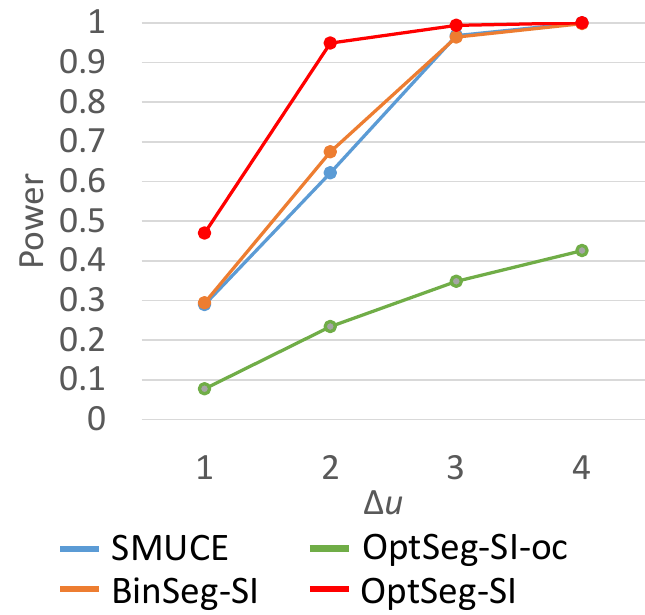}  
  \caption{Power}
  \label{fig:fixed_k_sub_second}
\end{subfigure}
\caption{False positive rate (FPR) and power comparison when $K$ is fixed.}
\label{fig:fixed_k}
\end{minipage}\hfill
\begin{minipage}{0.49\textwidth}
\begin{subfigure}{.495\linewidth}
  \centering
  \includegraphics[width=\linewidth]{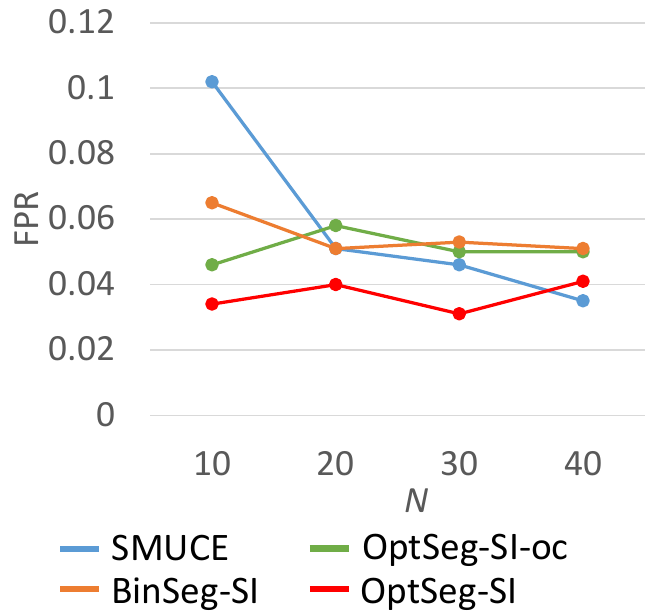}  
  \caption{False Positive Rate}
  \label{fig:adaptive_sub_first}
\end{subfigure}
\begin{subfigure}{.495\linewidth}
  \centering
  \includegraphics[width=\linewidth]{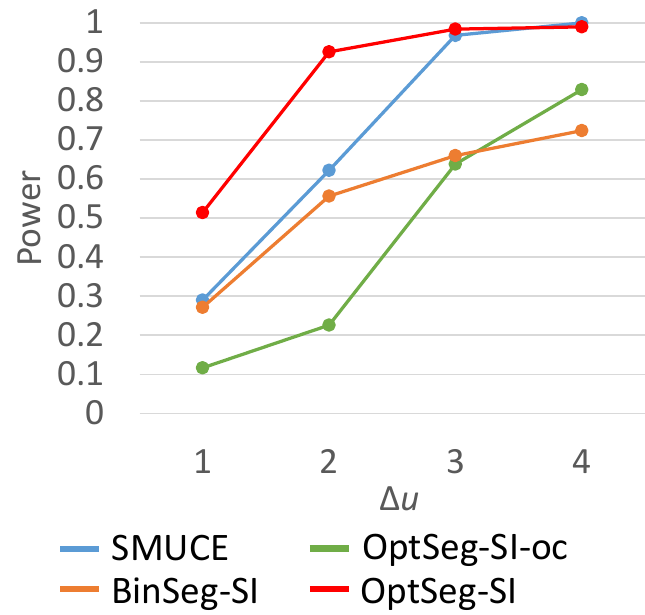}  
  \caption{Power}
  \label{fig:adaptive_sub_second}
\end{subfigure}
\caption{False positive rate (FPR) and power comparison when $K$ is unknown.}
\label{fig:adaptive}
\end{minipage}
\end{figure}

\begin{figure}[t]
\begin{minipage}{0.63\textwidth}
\centering
\includegraphics[width=0.8\linewidth]{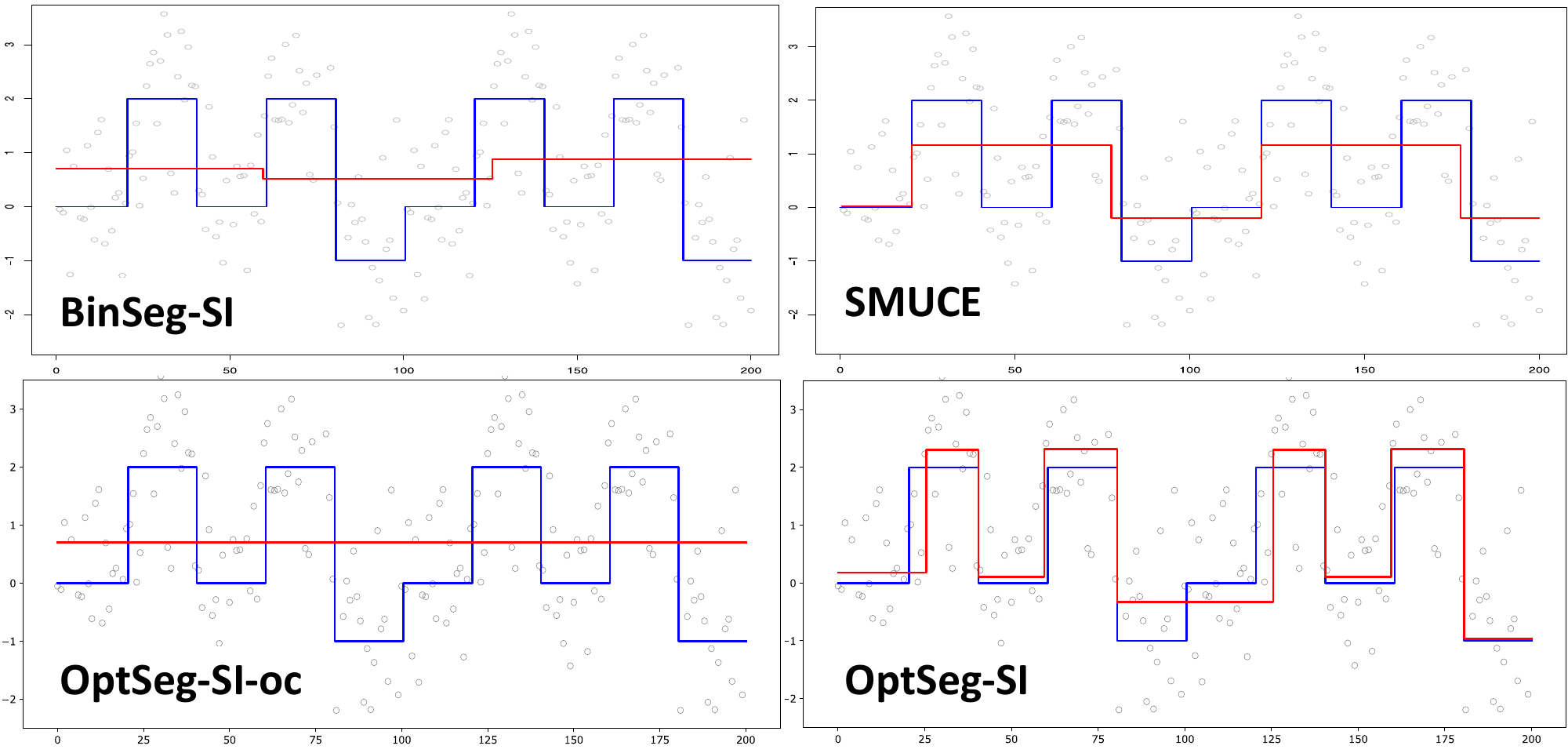}  
\caption{
Power demonstration of the OptSeg-SI method. The underlying mechanism (blue), data points (grey), and the results of each method (red) are shown in each panel. The result of OptSeg-SI is mostly close to the ground truth compared to the other methods.}
\label{fig:demonstration_ex1}
\end{minipage}\hfill
\begin{minipage}{0.35\textwidth}
\centering
\includegraphics[width=\linewidth]{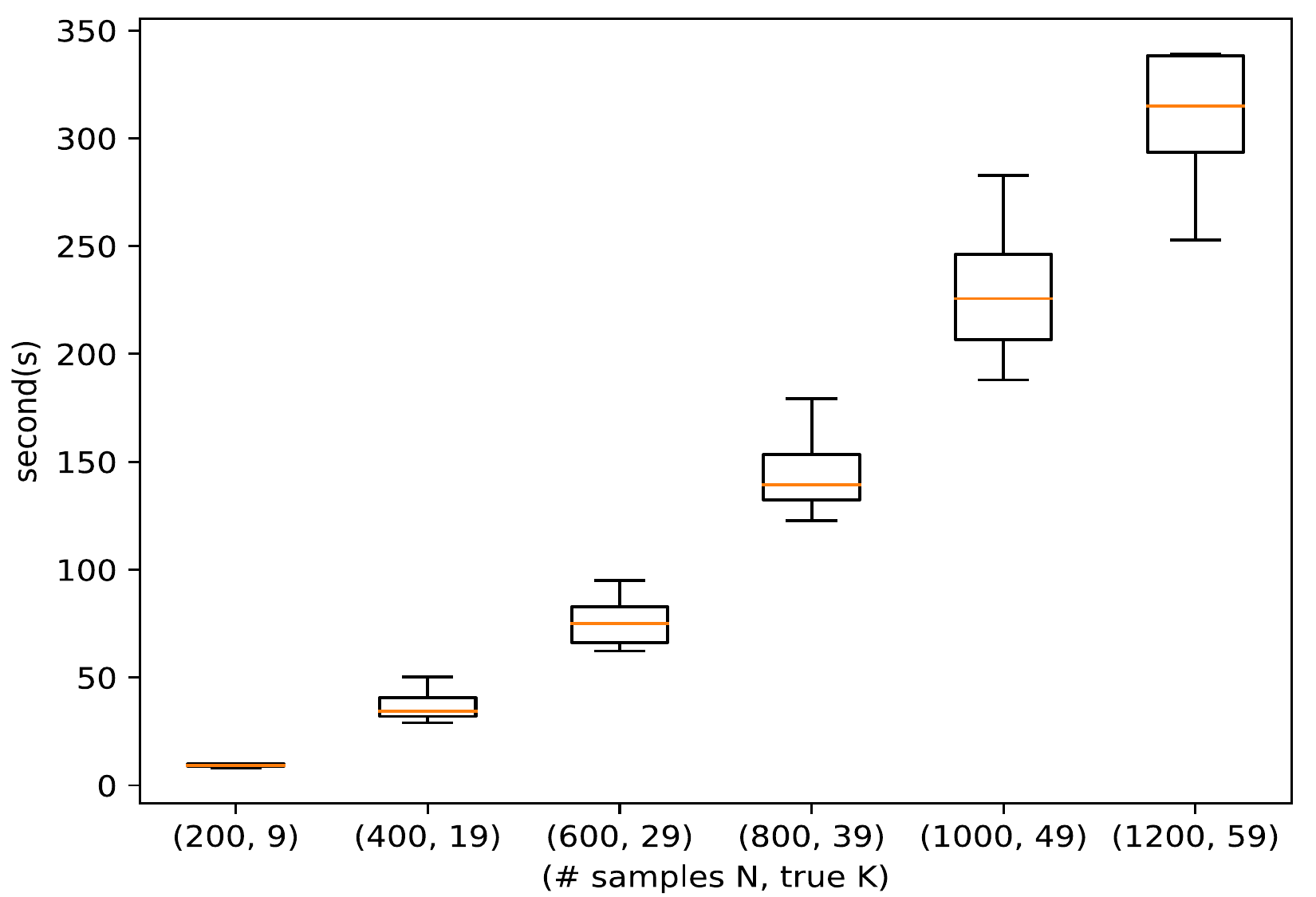}  
\caption{
Computing time of the OptSeg-SI method. The computing time of our proposed method is almost linear.}
\label{fig:computing_time}
\end{minipage}
\end{figure}

\textbf{Experimental results.} Figures \ref{fig:fixed_k} and \ref{fig:adaptive} respectively show the comparison results of the false positive rate (FPR) and true positive rate (TPR) when $K$ is fixed and $K$ is unknown.
In both cases, since SMUCE guarantee is only asymptotic, it could not control the FPR when $N$ is small.
While BinSeg-SI and OptSeg-SI-oc properly control the FPR, their powers are low because of over-conditioning.
OptSeg-SI always has high power while properly controlling the FPR.
Figure \ref{fig:demonstration_ex1} shows the power demonstration of the OptSeg-SI method.
While the existing methods missed many of true CPs, our method could identify almost all of them.
Figure \ref{fig:computing_time} shows the efficiency of OptSeg-SI method.% 
We generated data for each case $(N, K) \in  \{(200, 9), ..., (1200, 59)\}$. 
We ran 10 trials for each case. 

%============================

Besides, we also conducted the following experiments to demonstrate the robustness of the OptSeg-SI method in terms of the FPR control:

$\quad \bullet$ Non-normal data: we consider the data following Laplace distribution, skew normal distribution (skewness coefficient 10) and $t_{20}$ distribution.
In each experiment, we generated 12,000 null sequences for $N \in \{10, 20, 30, 40\}$.
We test the FPR for both $\alpha = 0.05$ and $\alpha = 0.1$. 
We confirmed that our method still maintains good performance on FPR control.
% under significance level $\alpha$. 
The results are shown in Appendix \ref{appendix:numerical_experiments}.
Besides, for dealing with the case of non-normal data, we can also apply a popular Box-Cox transformation \cite{box1964analysis} to the data before performing our method.\\ 
$~\quad \bullet$ Unknown $\sigma^2$: we consider the case when the variance is also estimated from the data. 
We generated 12,000 null sequences for $N \in \{50, 60, 70, 80\}$.
Our OptSeg-SI method still can properly control the FPR. The results are shown in Appendix \ref{appendix:numerical_experiments}.

We also performed TPR comparison on real-world dataset in which we showed that our method always has higher power compared to other existing method. 
We used \emph{jointseg} package \cite{pierre2014performance} to generate realistic DNA copy number profiles of cancer samples with ``known" truth.
Two datasets with 1,000 profiles of length $N=60$ and true $K = 2$ for each were created as follows: 

$\quad \bullet$ $\bm{\cD}_1$: Resample from GSE11976 with tumor fraction $=$ 1\\
$~\quad \bullet$ $\bm{\cD}_2$: Resample from GSE29172 with tumor fraction $=$ 1

The results are shown in Table \ref{tbl:power_comparison_real_data}.
Our proposed OptSeg-SI has higher power than the other methods in all cases.
We also applied OptSeg-SI to the Array CGH data provided by \citet{snijders2001assembly} and the Nile data which contains annual flow volume of the Nile river. %at Aswan from 1871 to 1970.
All of the results are consistent with \citet{snijders2001assembly, jung2017bayesian}. More details of the results can be found in Appendix \ref{appendix:real_experiments}.

\begin{table}[!t]
\vspace{-10pt}
\centering
\renewcommand{\arraystretch}{1.2}
\caption{Power comparison on real-world bioinformatics related datasets.}
\label{tbl:power_comparison_real_data}
{\footnotesize
\begin{tabular}{|l|c|c|c|c|}
\hline
  & SMUCE & OptSeg-SI-oc & BinSeg-SI & OptSeg-SI \\ 
\hline
  $\bm{\cD}_1$ &  0.53 & 0.24  & 0.33 &  \textbf{0.75} \\ 
\hline
  $\bm{\cD}_2$  & 0.62 & 0.27 &  0.32 & \textbf{0.71} \\ 
\hline
\end{tabular}
}
\vspace{-10pt}
\end{table}

%\section{Discussion}
%We investigate the robustness of our proposed method
%%when some assumptions are violated 
%in terms of the FPR control.
%
%$\bullet$ Non-normal data: we consider the data following Laplace distribution, skew normal distribution (skewness coefficient 10) and $t_{20}$ distribution.
%%
%In each experiment, we generated 12,000 null data for $N \in \{10, 20, 30, 40\}$.
%%
%We test the FPR for both $\alpha = 0.05$ and $\alpha = 0.1$. 
%%
%We confirmed that our method still maintains good performance on FPR control.
%% under significance level $\alpha$. 
%The results are shown in Appendix \ref{appendix:violate_assumption}.
%
%Besides, for dealing with the case of non-normal data, we can also apply a popular Box-cox transformation \cite{box1964analysis} to the data before performing our method.
%
%$\bullet$ Unknown $\sigma^2$: we consider the case when the variance is directly estimated from the data. 
%%
%We generated 12,000 null data for $N \in \{50, 60, 70, 80\}$ and conducted experiments.
%%
%Our proposed method still can properly control the FPR. The results are shown in Appendix \ref{appendix:violate_assumption}.

\section{Conclusion}

In this paper, we have introduced a powerful SI approach for the CP detection problem. 
We have conducted experiments on both synthetic and real-world datasets to show the good performance of the proposed OptSeg-SI method.
In the future, we could extend our method to the case of multi-dimensional sequences \cite{umezu2017selective}. For this case, computational efficiency is also a big challenge. Therefore, providing an efficient approach would also represent a valuable contribution.

\section*{Broader Impact}
Reliable machine learning (ML), which is the problem of assessing the reliability of data-driven knowledge obtained by ML algorithms, is one of the most important issues in the ML community.
Changepoint (CP) detection is an important unsupervised learning task, and has been studied in many areas.
Unfortunately, less attention has been paid to the statistical reliability of the detected CPs. 
Without statistical reliability, the results may contain many \emph{false detections}. 
These falsely detected CPs are harmful when they are used for high-stake decision making.

The main idea of this paper is to employ a selective inference --- a new promising approach for assessing the statistical reliability of data-driven hypotheses selected by complex data analysis algorithms --- to quantify the reliability of the detected CPs.
By mainly focusing on the reliability, this paper can have potential impact on 
reducing the risky as well as improving the quality of several CP detection-based data analysis tasks such as bioinformatics \cite{frick2014multiscale, pierre2014performance}, financial analysis \cite{fryzlewicz2014wild}, climatology \cite{killick2012optimal}, signal processing \cite{jandhyala2013inference}.
Especially for applications in healthcare domain, since the $p$-value that we introduced in the paper is valid and it is guaranteed that the probability of making false decisions is properly controlled, valid $p$-values can be used as one of many other possible criteria for making medical decisions.

\begin{ack}
This work was partially supported by MEXT KAKENHI (20H00601, 16H06538), JST CREST (JPMJCR1502), RIKEN Center for Advanced Intelligence Project, and RIKEN Junior Research Associate Program.
\end{ack}

\medskip

\bibliographystyle{abbrvnat}
\bibliography{ref}

\clearpage
\appendix
\section{Appendix} 

\subsection{Proof for Lemma 1}  \label{appendix:proof_lemma_1}
\noindent{\bf Lemma 1}.
{\it
For $n \in [N]$ and $k \in [K]$,
the set of CP vectors 
having potential to be optimal is
constructed as
\begin{align}
 \label{appendix:eq:T_Bellman}
 \hat{\cT}_{k, n} = {\mathop{\cup}}_{m=k}^{n-1} \{{\tt concat}({\cT}^{\rm opt}_{k-1, m}, m)\},
\end{align}
where
we extend the 
${\tt concat}$
operator
for the case where the first argument is a set of vectors,
which simply returns the set of concatenated vectors. 
}

\noindent{\it Proof.}
We prove the lemma by showing that
any CP vector
$\bm \tau \not \in {\cT}^{\rm opt}_{k-1, m}$,
for
$m \in \{k, \ldots, n-1\}$, 
cannot be subvector of the optimal CP vectors for problems with larger 
$n$ and $k$ 
for 
any
$z \in \RR$,
i.e., ${\tt concat}(\bm \tau, m) \not \in {\cT}^{\rm opt}_{k, n} $ for $n > m$.
For
$m \in \{k, \ldots, n-1\}$,
let
$\bm \tau \not \in {\cT}^{\rm opt}_{k-1, m}$
be a CP vector
which is NOT optimal for all 
$z \in \RR$,
i.e.,
\begin{align*}
 L_{k-1, m}(z, \bm \tau) > L^{\rm opt}_{k-1, m}(z)
 ~~~
 \forall z \in \RR.
\end{align*}
It suggests that,
for any
$m \in \{k, \ldots, n-1\}$
and
$z \in \RR$, 
\begin{align*}
 L^{\rm opt}_{k, n}(z) 
 &=
 \min_{m^\prime \in \{k, \ldots, n-1\}}
 \left(
 L^{\rm opt}_{k-1, m^\prime}(z)
 +
 C(\bm x(z)_{m^\prime + 1:n})
 \right)
 \\
 &
 \le
 L^{\rm opt}_{k-1, m}(z) +  C(\bm x(z)_{m+1:n})
 \\
 &
 <
 L_{k-1, m}(z, \bm \tau) + C(\bm x(z)_{m+1:n})
\end{align*}
for all $z \in \RR$. 
Thus, 
for any choice of $m \in \{k, \ldots, n-1\}$ and $z \in \RR$, 
$\bm \tau \not \in {\cT}^{\rm opt}_{k-1, m}$
cannot be a subvector of the optimal CP vector
for problems with larger $n$ and $k$.
In other words, 
only the CP vectors in
$\cup_{m=k}^{n-1} {\cT}^{\rm opt}_{k-1, m}$
can be used as the subvector of optimal CP vectors for problems with larger $n$ and $k$. 

%============================

\subsection{Proofs for Lemma 2 and 3 for the case when $K$ is unknown in \S 4}  \label{appendix:proof_lemma_2_3}
\noindent{\bf Lemma 2}.
{\it
For $m < n$, if a vector $\bm \tau \not \in \cT^{\rm opt}_m$, then ${\tt concat}(\bm \tau, m) \not \in \cT^{\rm opt}_n$.
}

\noindent{\it Proof.}
For $m < n$, if a vector $\bm \tau \not \in \cT^{\rm opt}_m$, 
\begin{align*}
	L_m(z, \bm \tau) > L^{\rm opt}_m(z) \quad \forall z \in \RR.
\end{align*}
It suggests that, for any $m \in \{0, ..., n -1\}$ and $z \in \RR$,
\begin{align*}
	L^{\rm opt}_n(z) &= \min \limits_{m^\prime \in \{0, ..., n-1\}} \{L^{\rm opt}_{m^\prime}(z) + C(\bm x(z)_{m^\prime + 1:n}) + \beta \} \\
	& \leq L^{\rm opt}_{m}(z) + C(\bm x(z)_{m + 1:n}) + \beta \\	
	& < L_m(z, \bm \tau) + C(\bm x(z)_{m + 1:n}) + \beta.
\end{align*}
Therefore, for any $m \in \{0, ..., n-1\}$, if $\bm \tau \not \in \cT^{\rm opt}_m$, then ${\tt concat}(\bm \tau, m) \not \in \cT^{\rm opt}_n$. 

%============================
\noindent{\bf Lemma 3}.
{\it
For $m < n$, if $\bm \tau \not \in \cT^{\rm opt}_m$ and 
\begin{align*}
	L_{m}(z, \bm \tau) - \beta > L^{\rm opt}_m(z) \quad \forall z \in \RR
\end{align*}
holds, then $\bm \tau \not \in \cT^{\rm opt}_n$.
}

\noindent{\it Proof.} 
For any $m \in \{0, ..., n -1\}$ and $z \in \RR$, we have 
\begin{align*}
	L^{\rm opt}_n(z) &= \min \limits_{m^\prime \in \{0, ..., n-1\}} \{L^{\rm opt}_{m^\prime}(z) + C(\bm x(z)_{m^\prime + 1:n}) + \beta \} \\
	& \leq L^{\rm opt}_{m}(z) + C(\bm x(z)_{m + 1:n}) + \beta.
\end{align*}
For any $m \in \{0, ..., n -1\}$, if a CP vector $\bm \tau \not \in \cT^{\rm opt}_m$ satisfies Lemma 3, then it suggests
\begin{align*}
	L^{\rm opt}_n(z) & \leq L^{\rm opt}_{m}(z) + C(\bm x(z)_{m + 1:n}) + \beta \\
	\Leftrightarrow \quad L^{\rm opt}_n(z) &< L_{m}(z, \bm \tau) - \beta + C(\bm x(z)_{m + 1:n}) + \beta \\
	\Leftrightarrow \quad L^{\rm opt}_n(z) &< L_{m}(z, \bm \tau) + C(\bm x(z)_{m + 1:n})  
\end{align*}
for all $z \in \RR$. On the other hand, we have 
\begin{align*}
  L_{m}(z, \bm \tau) + C(\bm x(z)_{m + 1:n})  \leq L_{n}(z, \bm \tau)
\end{align*}
holds for any $z \in \RR$ because the cost is always reduced when adding a changepoint at position $m$ without the penalty term. Hence, we have
\begin{align*}
	L^{\rm opt}_n(z) < L_{n}(z, \bm \tau)
\end{align*}
for all $z \in \RR$. Therefore, $\bm \tau \not \in \cT^{\rm opt}_n$ and Lemma 3 holds.

%============================
\subsection{Additional tricks for methods proposed in \S3.} \label{appendix:computational_trick}
\paragraph{Finding optimal CP vector when $z = - \infty$ in {\tt paraCP}($n, k, \hat{\cT}_{k, n}$) in Algorithm \ref{alg:paraCP}.}
For each $\bm \tau \in \hat{\cT}_{k, n}$,  the corresponding loss function at $\bm \tau$ is written as a positive definite quadratic function.
Therefore, at $z = - \infty$, the optimal CP vector is the one whose corresponding loss function $L_n(z, \bm \tau)$ has the smallest coefficient of the quadratic term. 
If there are more than one quadratic function having the same smallest quadratic coefficient, we then choose the one that has the largest coefficient in the linear term. 
If those quadratic functions still have the same largest linear coefficient, we finally choose the one that has the smallest constant term.

\paragraph{Additional pruning condition for parametric DP when $K$ is fixed.}
In \S3.3, we showed that $\cT^{\rm opt}_{k, n}$ can be constructed from the set $\hat{\cT}_{k, n} \subseteq \cT_{k, n}$.
%However, for this additional pruning condition, we prove that $\cT^{\rm opt}_{k, n}$ can be constructed from a set $\bar{\cT}_{k, n}$ that is smaller than $\hat{\cT}_{k, n}$ by using the following Lemma.
%
By using the following lemma, we can construct a smaller superset of $\cT^{\rm opt}_{k, n}$, which leads to further efficiency of parametric DP.

\begin{lemma} \label{app:addition_lemma}
For $n \in [N]$, and $k \in [K]$, %the set $\bar{\cT}_{k, n}$ can be constructed by
let
\begin{equation*}
	\bar{\cT}_{k, n} = \{\bmtau \in \hat{\cT}_{k, n - 1} \setminus P_{\rm prune} \} \ \cup\ \{{\rm concat}(\cT^{\rm opt}_{k - 1, n - 1}, n - 1)\},
\end{equation*}
where
\begin{equation*}
P_{\rm prune} = \{\bmtau \in \hat{\cT}_{k, n - 1} \mid L_{k, n - 1} (z, \bmtau) > L^{\rm opt}_{k - 1, n - 1}(z), \forall z \in \mathbb{R} \}.
\end{equation*}
Then $\cT^{\rm opt}_{k, n} \subseteq \bar{\cT}_{k, n} \subseteq \hat{\cT}_{k, n}$.
\end{lemma}

\noindent{\it Proof.} First, to show $\hat{\cT}_{k, n} \supseteq \bar{\cT}_{k, n}$, from \eq{appendix:eq:T_Bellman},
\begin{align*}
	\hat{\cT}_{k, n} &= {\mathop{\cup}}_{m=k}^{n-1} \{{\tt concat}({\cT}^{\rm opt}_{k-1, m}, m)\}\\
		&= {\mathop{\cup}}_{m=k}^{n-2} \{{\tt concat}({\cT}^{\rm opt}_{k-1, m}, m)\} \mathop{\cup} \{{\tt concat}({\cT}^{\rm opt}_{k-1, n-1}, n-1)\}\\
		&= \hat{\cT}_{k, n - 1} \mathop{\cup} \{{\tt concat}({\cT}^{\rm opt}_{k-1, n-1}, n-1)\}\\
		&\supseteq  \{ \hat{\cT}_{k, n - 1} \setminus P_{\rm prune}\} \mathop{\cup} \{{\tt concat}({\cT}^{\rm opt}_{k-1, n-1}, n-1)\} = \bar{\cT}_{k, n}.
\end{align*}
Next, to show $\cT^{\rm opt}_{k, n} \subseteq \bar{\cT}_{k, n}$, we only need to prove that $\bmtau \in P_{\rm prune}$ never be the optimal CP vector at $k, n$, i.e., $\bmtau \not \in \cT^{\rm opt}_{k, n}$. For any $\bm \tau \in P_{\rm prune}$
\begin{align*}
	L_{k, n}(z, \bm \tau) &\geq L_{k, n - 1}(z, \bm \tau)\\
	&> L^{\rm opt}_{k - 1, n - 1}(z)\\
	&=  L^{\rm opt}_{k - 1, n - 1}(z) + C(x(z)_{n:n})\\
	&\geq  \min \limits_{m^\prime \in \{k, ..., n -1\}} (L^{\rm opt}_{k - 1, m^\prime}(z) + C(x(z)_{(m^\prime + 1):n})) \\
	&= L^{\rm opt}_{k, n} (z),
\end{align*}
for any $z \in \RR$. Therefore, $\bmtau \in P_{\rm prune}$ never belongs to $\cT^{\rm opt}_{k, n}$.

%============================

\subsection{Distribution of naive $p$-value and selective $p$-value when the null hypothesis is true} \label{appendix:distribution_p_value}
\begin{figure}[H]
\begin{subfigure}{0.245\linewidth}
\centering
\includegraphics[width=\linewidth]{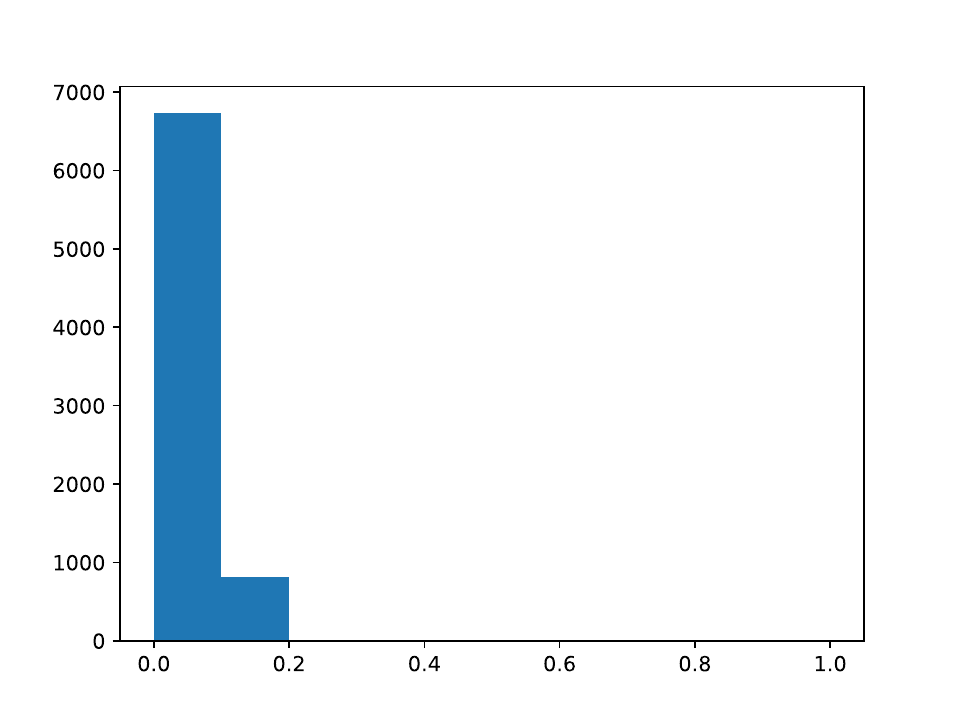}
\caption{$N = 10$}
\end{subfigure}
\begin{subfigure}{0.245\linewidth}
\centering
\includegraphics[width=\linewidth]{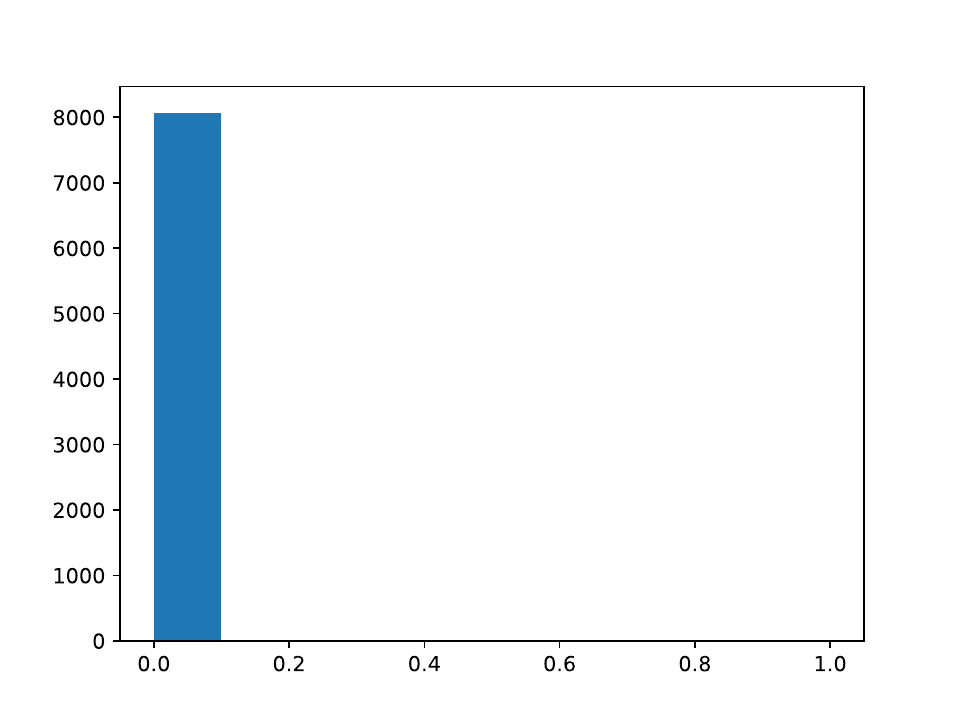}
\caption{$N = 20$}
\end{subfigure}
\begin{subfigure}{0.245\linewidth}
\centering
\includegraphics[width=\linewidth]{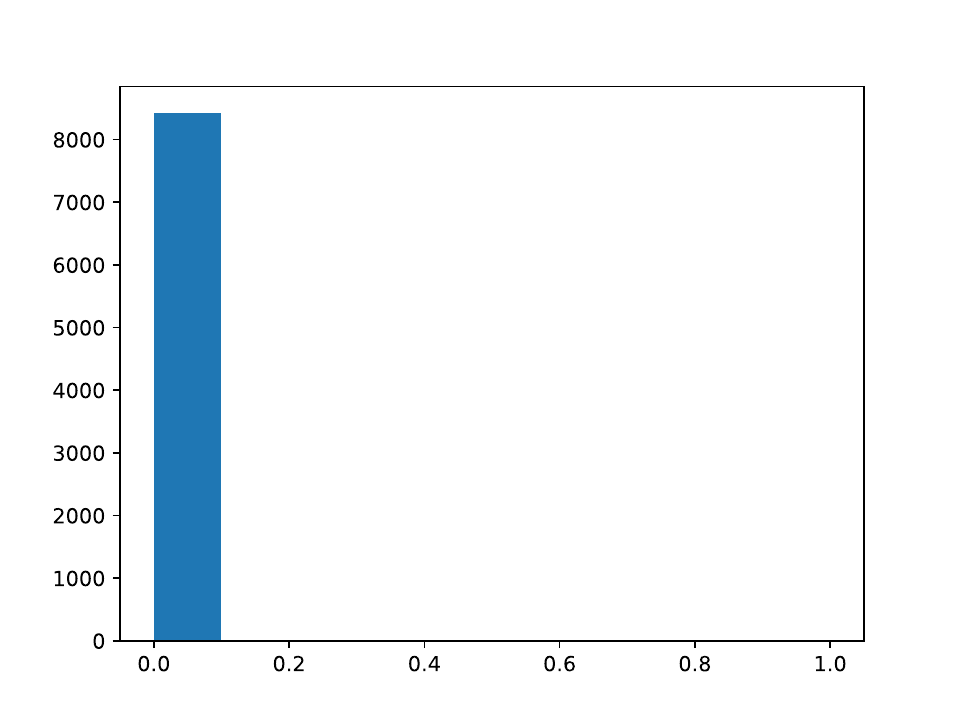}
\caption{$N = 30$}
\end{subfigure}
\begin{subfigure}{0.245\linewidth}
\centering
\includegraphics[width=\linewidth]{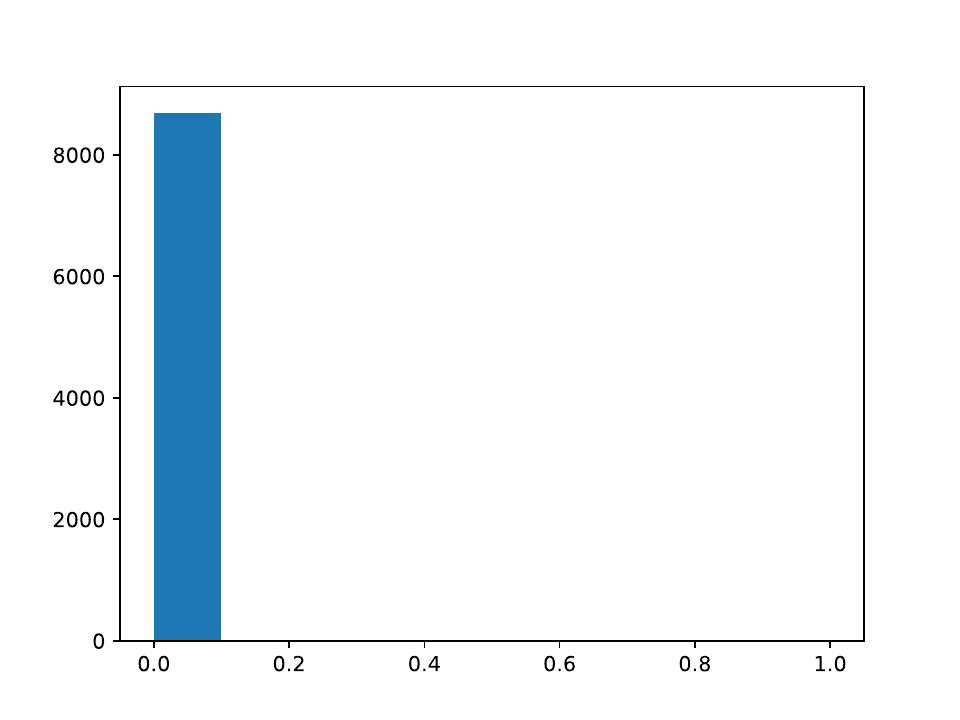}
\caption{$N = 40$}
\end{subfigure}
\caption{Distribution of naive $p$-value when the null hypothesis is true.}
\label{fig:naive_p_value_distribution}
\end{figure}

\begin{figure}[H]
\begin{subfigure}{0.245\linewidth}
\centering
\includegraphics[width=\linewidth]{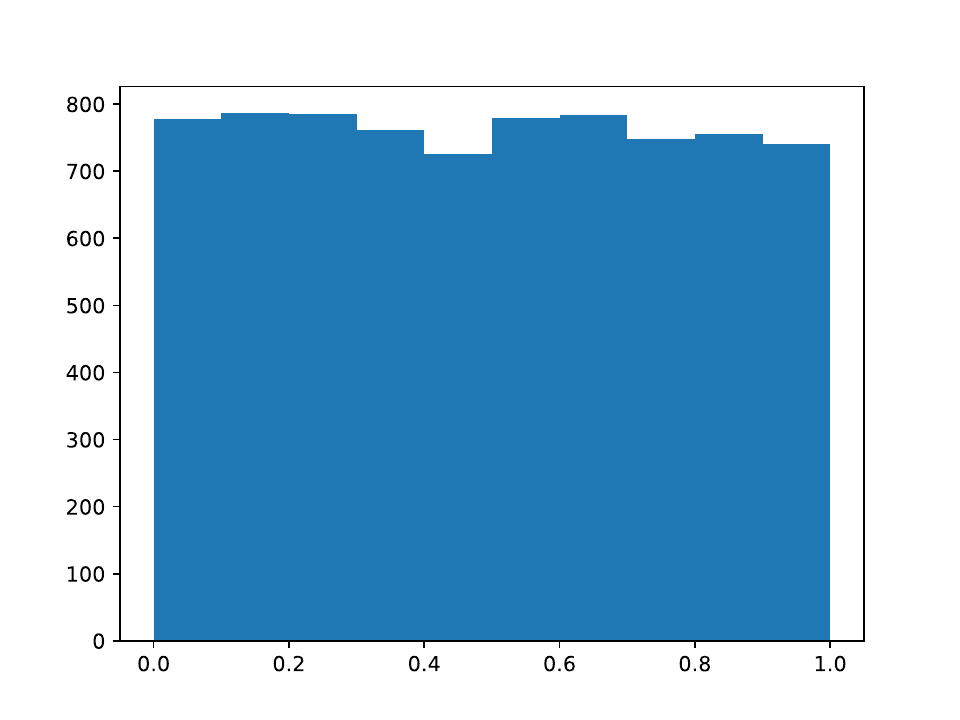}
\caption{$N = 10$}
\end{subfigure}
\begin{subfigure}{0.245\linewidth}
\centering
\includegraphics[width=\linewidth]{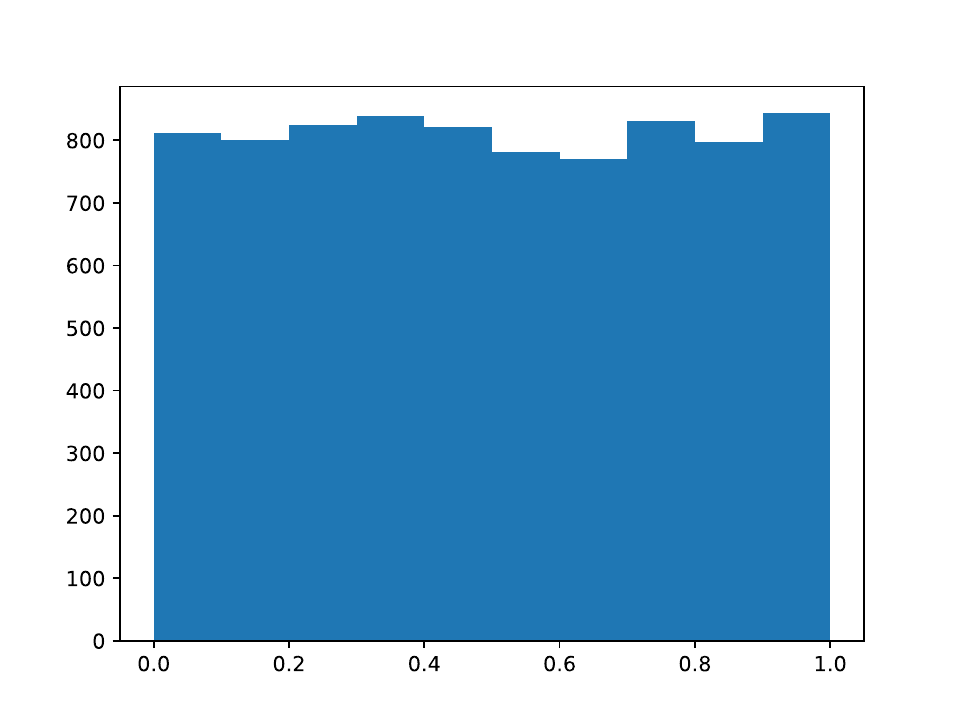}
\caption{$N = 20$}
\end{subfigure}
\begin{subfigure}{0.245\linewidth}
\centering
\includegraphics[width=\linewidth]{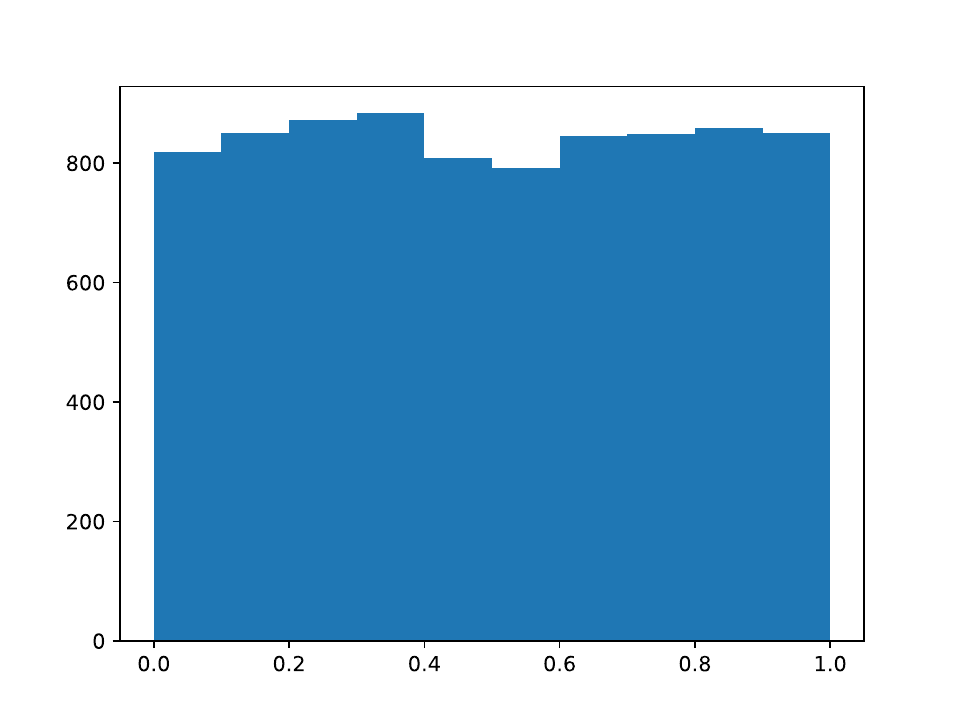}
\caption{$N = 30$}
\end{subfigure}
\begin{subfigure}{0.245\linewidth}
\centering
\includegraphics[width=\linewidth]{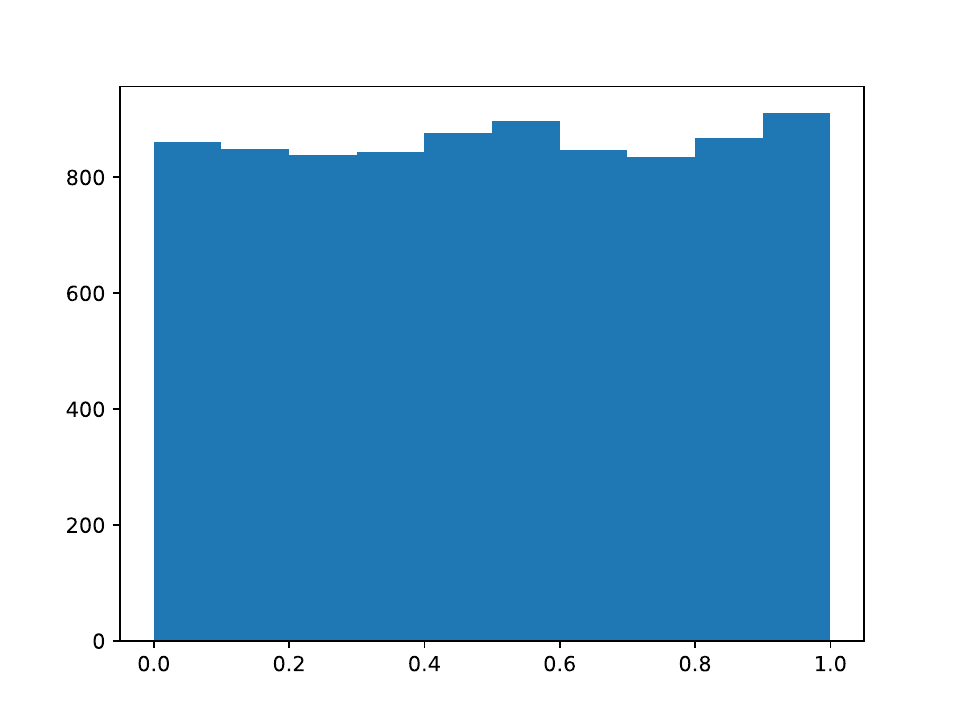}
\caption{$N = 40$}
\end{subfigure}
\caption{Distribution of selective $p$-value when the null hypothesis is true.}
\label{fig:selective_p_value_distribution}
\end{figure}

We demonstrate the \emph{validity} of our proposed OptSeg-SI method by confirming the uniformity of $p$-value when the null hypothesis is true. 
We generated 12,000 null sequences $\bmx = (x_1, ..., x_N)$ in which $x_{i \in [N]} \sim \NN(0, 1)$ for each case $N \in \{10, 20, 30, 40\}$ and performed the experiments to check the distribution of naive $p$-values and selective $p$-values.
From Figure \ref{fig:naive_p_value_distribution}, it is obvious that naive $p$-value does not follow uniform distribution. Therefore, it fails to control the false positive rate.
The empirical distributions of selective $p$-value are shown in Figure \ref{fig:selective_p_value_distribution}.
The results indicate our proposed method successfully control the false detection probability.

%============================

%============================
\subsection{Details for numerical experiments.} \label{appendix:numerical_experiments}

\paragraph{Methods for Comparison.}
We compared the performance of the OptSeg-SI with the following approaches:

$\bullet$ \textbf{SMUCE~\cite{frick2014multiscale}.}
This is asymptotic test for multiple detected CPs. %SMUCE was also used for comparison in \citet{hyun2018exact}. 
The implementation of SMUCE is available at \url{https://cran.r-project.org/web/packages/stepR/index.html}.

$\bullet$ \textbf{[BinSeg-SI] SI for Binary Segmentation~\cite{hyun2018post}}
%This is SI method for binary segmentation.
%
In \citet{hyun2018post}, it was reported that SI for Fused Lasso (proposed by the same authors), is worse than BinSeg-SI. Therefore, we only compare to BinSeg-SI. 
BinSeg-SI had been considered as a computationally efficient approximation of the problem in \eq{eq:selective_p}, where 
the authors additionally condition on extra information for computational tractability, e.g., the order that CPs are detected.
This is one of the reasons why BinSeg-SI has low power.
The implementation of BinSeg-SI is available at \url{https://github.com/robohyun66/binseginf}.

$\bullet$ \textbf{[OptSeg-SI-oc] SI method for optimal CPs with %suboptimal
 {\bf o}ver-{\bf c}onditioning.}
In SI, there are mainly two approaches to characterize the selection event.
In the first approach, the selection event is only constructed based on the optimality condition of the problem, which is usually difficult or computationally impractical.
Therefore, the second approach is used to overcome the computational challenge by additionally conditioning on extra event.
Although the type I error can be properly controlled in the second approach, the power is generally low %compared to the first approach 
because of \emph{over-conditioning}.

To see the advantage of minimum conditioning of the proposed method,
we compare with two variants of SI 
%methods 
for optimal CPs
(each for fixed $K$ and unknown $K$ cases), which we call \emph{OptSeg-SI-oc}. 
In each of these variants,
instead of the truncation region
$\cZ$ characterized in the main paper,
its subsets are used as the conditioning set.
These subsets are constructed
by considering all the operations
when DP algorithm is used
for detecting the optimal CPs.
The OptSeg-SI-oc method and BinSeg-SI in \citet{hyun2018post} are categorized as the second approach.
We actually first developed OptSeg-SI-oc as our first SI method for optimal CPs (unpublished). The derivation of OptSeg-SI-oc is shown in Appendix \ref{appendix:derive_dp_selection_event}. Then, its drawback (over-conditioning) was resolved by the proposed OptSeg-SI method in this paper.
%

%=======================

\begin{figure}[t]
\centering
\includegraphics[width=0.85\linewidth]{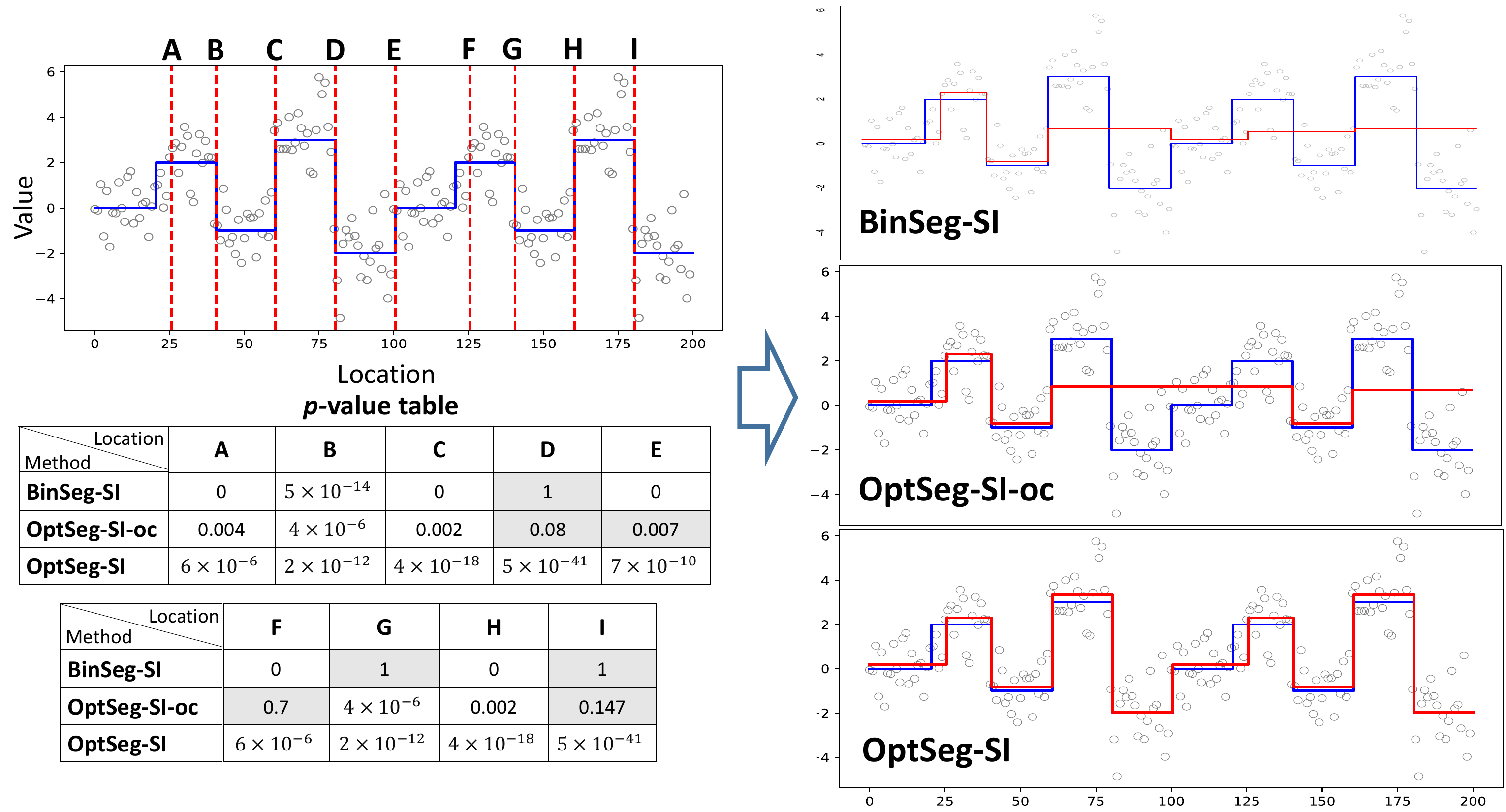}  
\caption{Additional results for power demonstration. 
In the left figure, the blue line and the grey circles indicate the underlying mean and the observed sequence, respectively.
The red dotted lines are the results of optimal segmentation (OptSeg) and binary segmentation (BinSeg) algorithms.
Here, the CP detection results of OptSeg and BinSeg were the same.
Then, the significance of each CP is tested.
With Bonferroni correction, to control false detection rate at $0.05$, the significance level is decided by $\frac{0.05}{9} \approx 0.006$.
Three different $p$-values are shown for each detected CP: BinSeg-SI $p$-value, OptSeg-SI-oc $p$-value and OptSeg-SI $p$-value.
BigSeg-SI missed many true CPs ({\bf D}, {\bf G}, {\bf I}). This problem is the same for OptSeg-SI-oc ({\bf D}, {\bf E}, {\bf F}, {\bf I}).
The OptSeg-SI method can identify all  true CPs.
The segments recovered based on the results of the significant testing from each method are shown in the right figure.
}
\label{fig:app_additional_demonstration}
\end{figure}

\paragraph{Experimental Results.} 
We show the detail of experimental results as follows:

$\bullet$ \textbf{Additional experiment for power demonstration of the proposed method.} In Figure \ref{fig:app_additional_demonstration}, we show additional results to demonstrate that our OptSeg-SI method can identify many true CPs.

\begin{figure}[t]
\centering
\begin{subfigure}{0.4\linewidth}
\centering
\includegraphics[width=\linewidth]{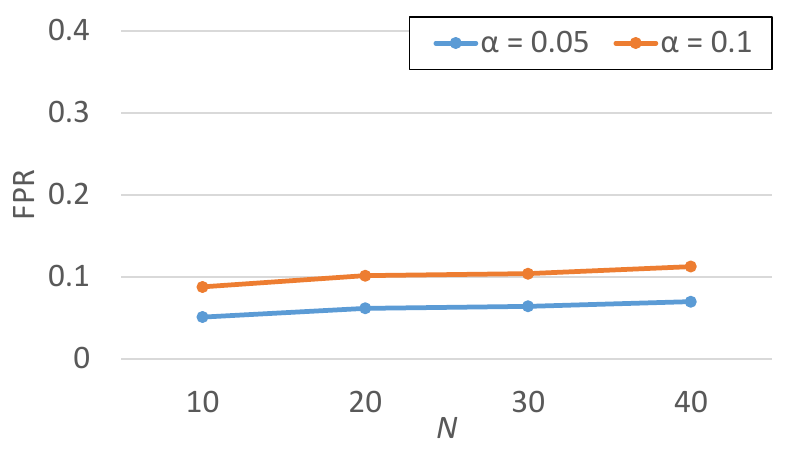}
\caption{Laplace distribution}
\label{fig:app_laplace}
\end{subfigure}
\begin{subfigure}{0.4\linewidth}
\centering
\includegraphics[width=\linewidth]{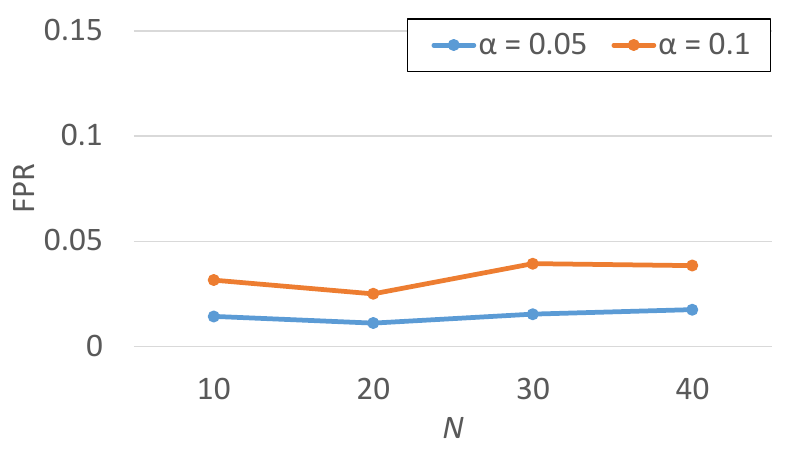}
\caption{Skew normal distribution}
\label{fig:app_skew_normal}
\end{subfigure}

\begin{subfigure}{0.4\linewidth}
\centering
\includegraphics[width=\linewidth]{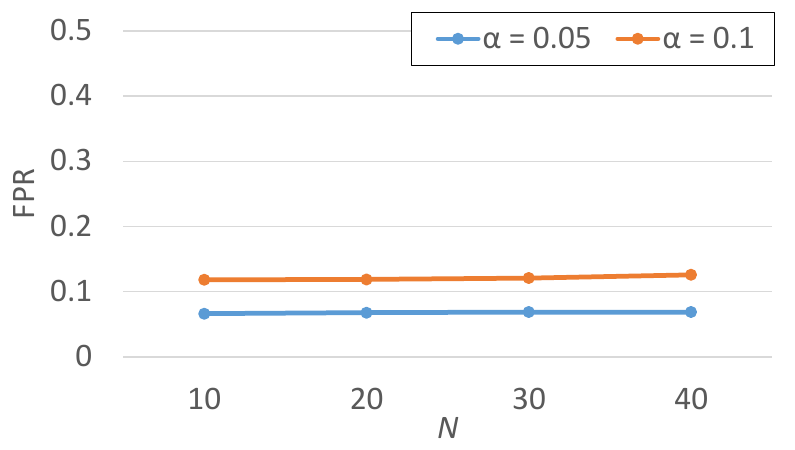}
\caption{$t_{20}$ distribution}
\label{fig:app_t_20}
\end{subfigure}
\begin{subfigure}{0.4\linewidth}
\centering
\includegraphics[width=\linewidth]{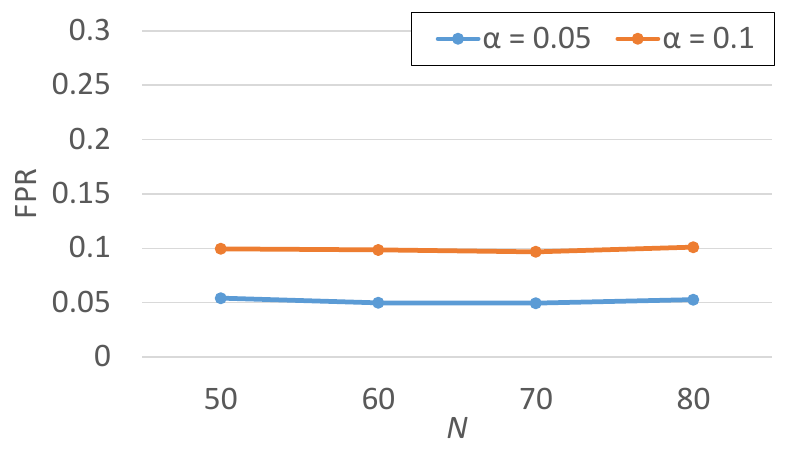}
\caption{$\sigma^2$ is unknown}
\label{fig:app_unknown_sigma}
\end{subfigure}
\caption{False positive rate of the proposed OptSeg-SI method when data is non-normal or $\sigma^2$ is unknown.}
\label{fig:violate_assumption}
\end{figure}

$\bullet$  \textbf{The robustness of the proposed OptSeg-SI method in terms of the FPR control.}
\begin{itemize}
\item[--]  Non-normal data: we considered the data following Laplace distribution, skew normal distribution (skewness coefficient 10) and $t_{20}$ distribution.
In each experiment, we generated 12,000 null sequences for $N \in \{10, 20, 30, 40\}$.
We tested the FPR for both $\alpha = 0.05$ and $\alpha = 0.1$. 
The FPR results are shown in Figure \ref{fig:app_laplace}, \ref{fig:app_skew_normal} and \ref{fig:app_t_20}. In case of Laplace distribution and skew normal distribution, our proposed method can properly control the FPR. For the case of $t_{20}$ distribution, the FPR is just a bit higher than the significance level.

\item[--] Unknown $\sigma^2$: 
We generated 12,000 null sequences $\bm x = (x_1, ..., x_N)$, in which $x_{i \in [N]} \sim \NN(0, 1)$, for $N \in \{50, 60, 70, 80\}$ and conducted experiments.
In this case, the value of $\sigma^2$ is also estimated from the data. We first perform CP detection algorithm to detect the segments. Since the estimated variance tends to be smaller than the true value, we calculated the empirical variance of each segment and set the maximum value for $\sigma^2$. The results are shown in Figure \ref{fig:app_unknown_sigma}.
Our proposed method still can properly control the FPR. 
\end{itemize}

\begin{figure}[t]
\centering
\includegraphics[width=0.45\linewidth]{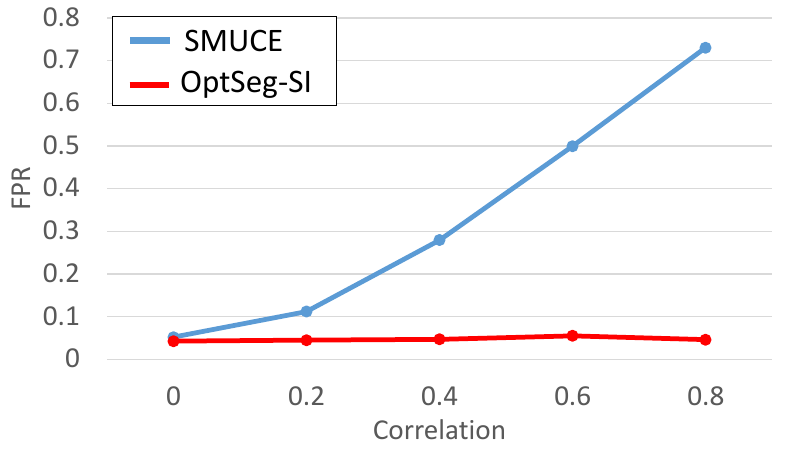}  
\caption{Comparison of FPR control when the sequence contains correlated data points.
With SMUCE, the FPR increases when the degree of correlation increases.
On the other hand, our proposed OptSeg-SI method can successfully control the FPR under $\alpha = 0.05$ in all cases.}
\label{fig:exp_fpr_dependent_data}
\end{figure}

$\bullet$ \textbf{Comparison of FPR control when the sequence contains correlated data points.}
In this experiment, we demonstrate that the asymptotic method (SMUCE) cannot control the FPR when the sequence contains correlated data points while our OptSeg-SI method can successfully control the FPR under the significance level $\alpha = 0.05$.
We generated 1,200 null sequences $\bm x = (x_1, ..., x_N) \sim \NN(\bm \mu, \bm \Xi)$, 
where $N = 20$, $\bm \mu = (\mu_1, ..., \mu_N)$ in which $\mu_{i \in [N]}=0$, 
and $\bm \Xi = \sigma^2 (\xi^{|i - j|})_{i,j \in [N]}$ in which $\xi$ is degree of correlation and $\sigma^2 = 1$.
We conducted experiments for $\xi \in \{0.0, 0.2, 0.4, 0.6, 0.8\}$.
The results are shown in Figure \ref{fig:exp_fpr_dependent_data}.
When $\xi = 0.0$, i.e., there is no correlation between the data points, SMUCE can control the FPR at $\alpha = 0.05$. However, when $\xi$ increases, the FPR also increases. It indicates that SMUCE cannot control the FPR when the data points are correlated.
On the other hand, our proposed OptSeg-SI method can successfully control the FPR under $\alpha$ in all cases.

%============================

\subsection{Details for real-data experiments.} \label{appendix:real_experiments}

\begin{figure}[t]
\begin{subfigure}{.49\linewidth}
  \centering
  \includegraphics[width=0.80\linewidth]{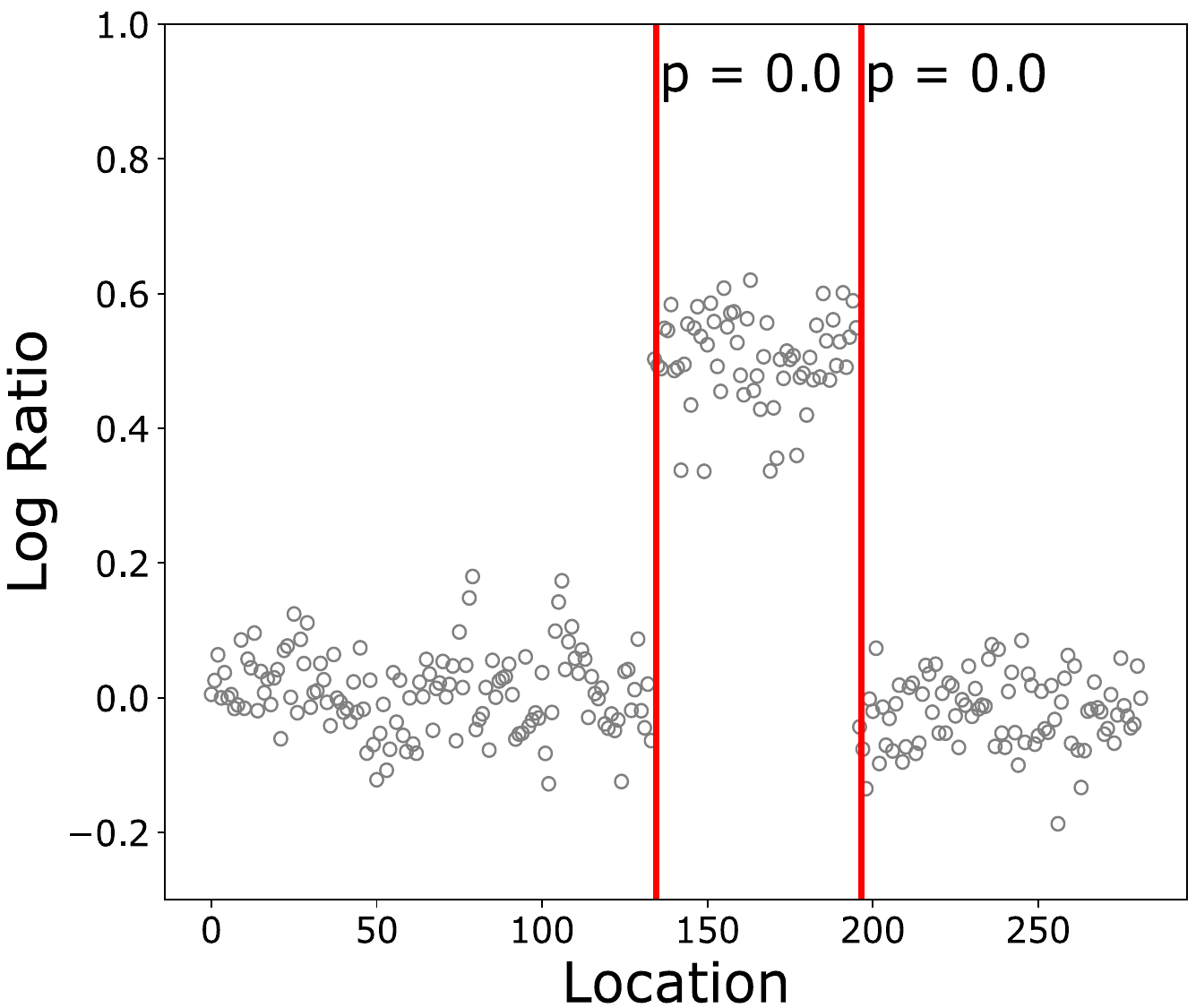}  
  \caption{Chromosomes 1, 2, 3.}
  \label{fig:sub_choromosomes_1_2_3}
\end{subfigure}
\begin{subfigure}{.49\linewidth}
  \centering
  \includegraphics[width=0.80\linewidth]{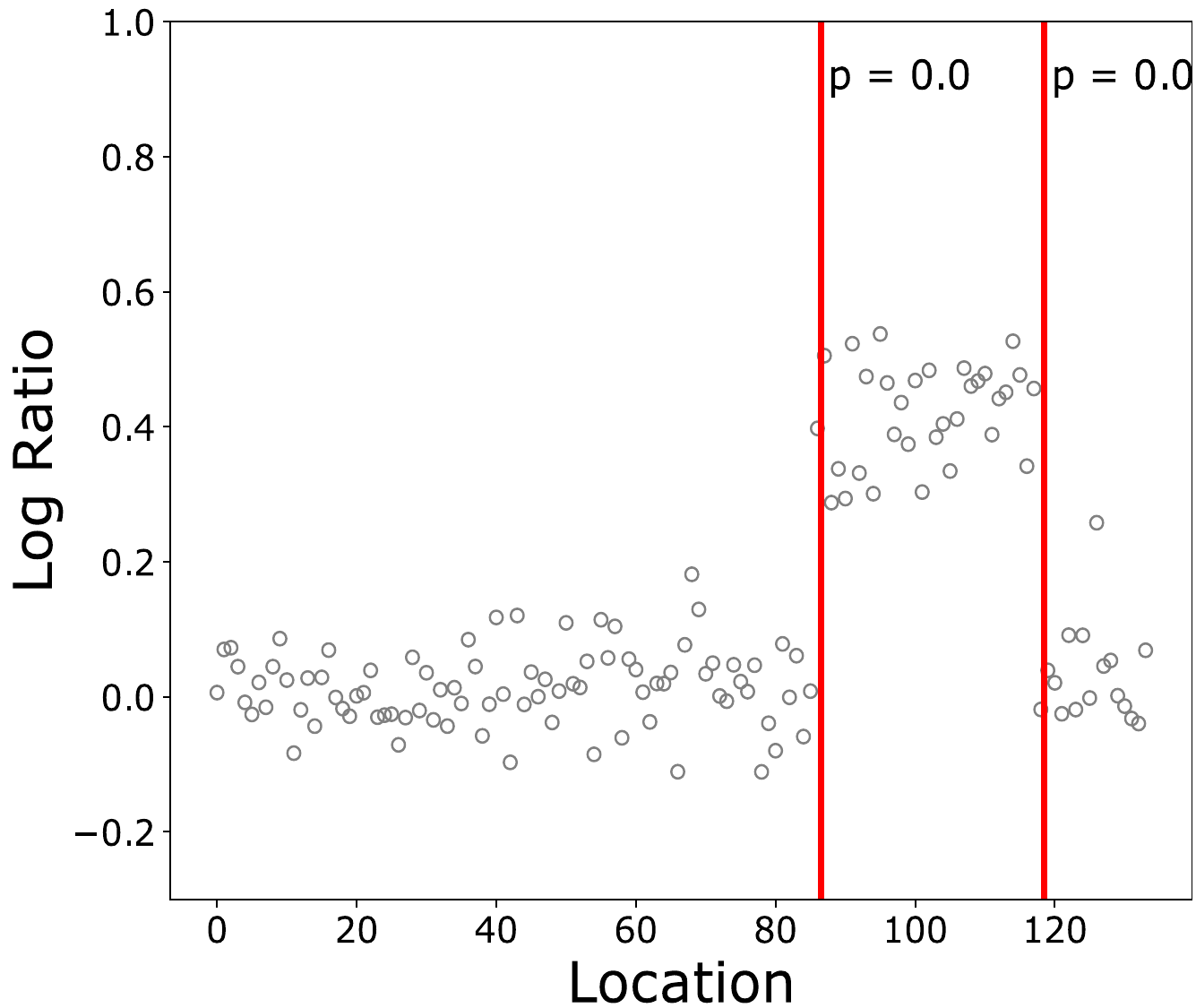}  
  \caption{Chromosomes 20, 21, 22.}
  \label{fig:sub_choromosomes_20_21_22}
\end{subfigure}
\caption{Experimental results for cell line GM03576.}
\label{fig:exp_cgh_GM03576}
\end{figure}

\begin{figure}[t]
\begin{subfigure}{.49\linewidth}
  \centering
  \includegraphics[width=0.8\linewidth]{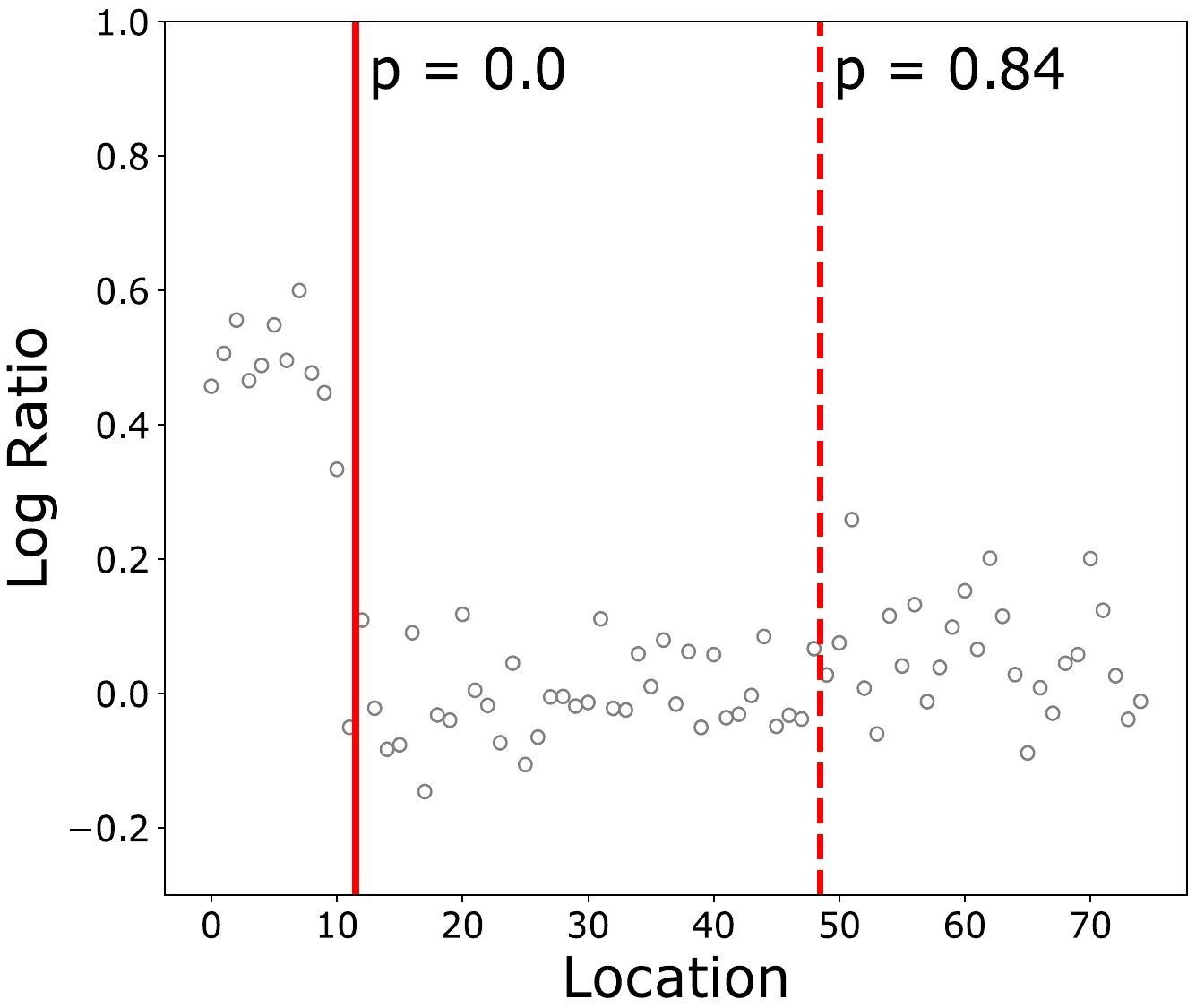}  
  \caption{Chromosome 14.}
  \label{fig:sub_chromosome_14}
\end{subfigure}
\begin{subfigure}{.49\linewidth}
  \centering
  \includegraphics[width=0.8\linewidth]{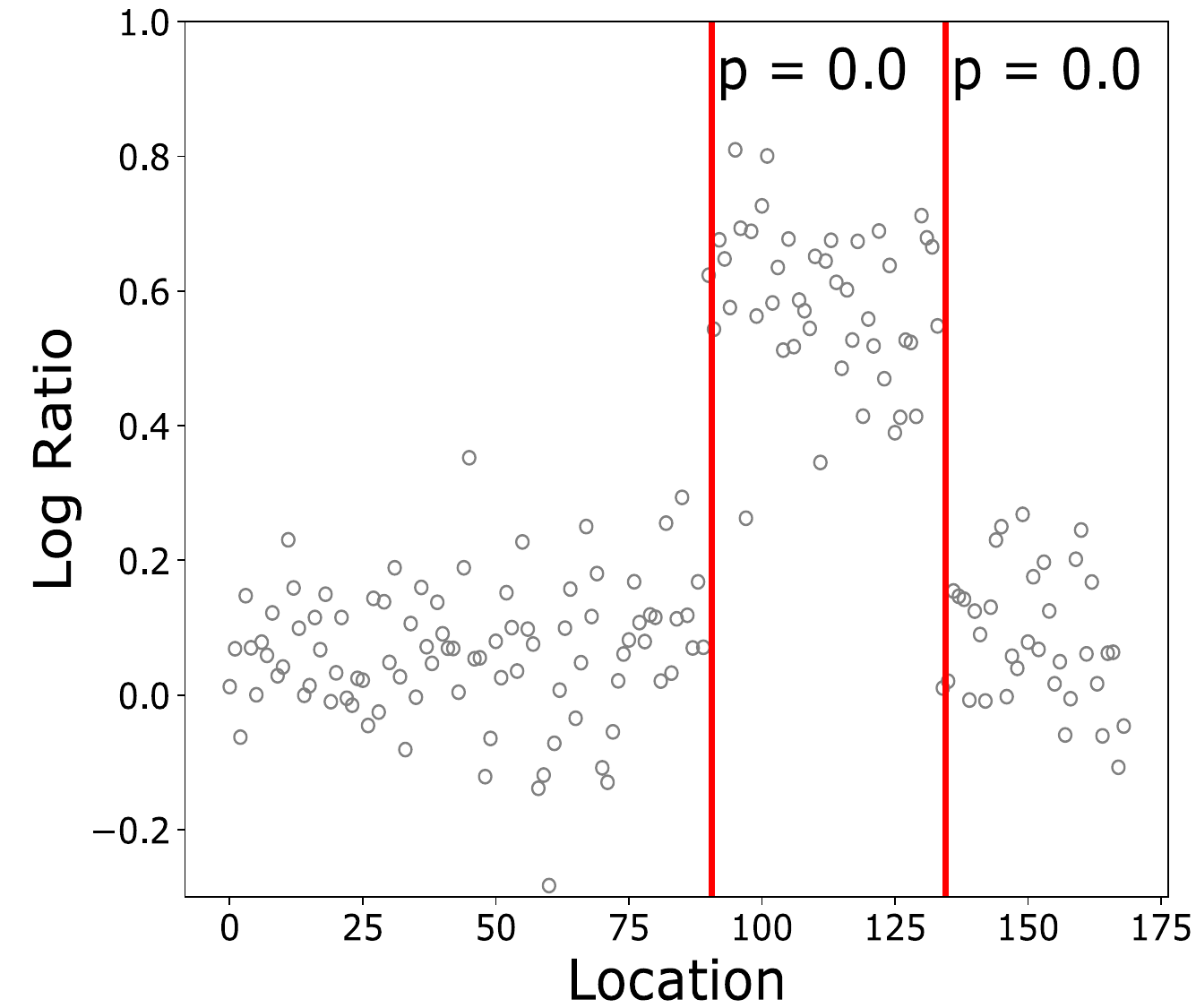}  
  \caption{Chromosomes 17, 18, 19}
  \label{fig:sub_choromosomes_17_18_19}
\end{subfigure}
\caption{Experimental results for cell lines GM00143 and GM01750.}
\label{fig:exp_cgh_GM00143_GM01750}
\end{figure}

\paragraph{Array CGH data.} Array CGH analyses detect changes in expression levels across the genome. The dataset with ground truth was provided in \citet{snijders2001assembly}. The results from our method were shown in Figure \ref{fig:exp_cgh_GM03576} and \ref{fig:exp_cgh_GM00143_GM01750}. The solid red line denotes the significant changepoint which has the $p$-value smaller than the significance level after Bonferroni correction. All of the results are consistent with \citet{snijders2001assembly}.

\begin{figure}[!t]
\centering
\includegraphics[width=0.55\linewidth]{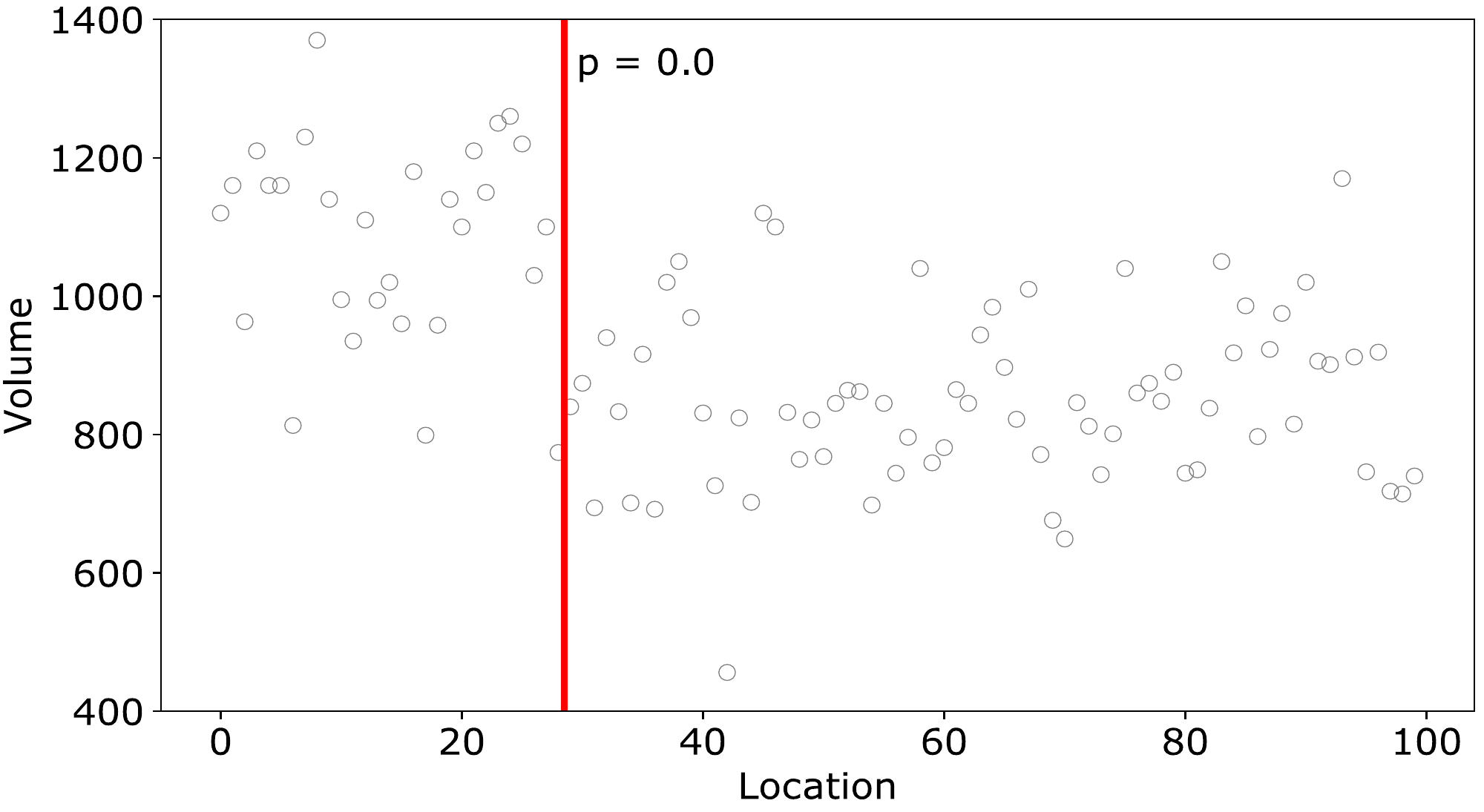}  
\caption{Experimental result for Nile data. The changepoint is detected at $28^{\rm th}$ position which indicates there is a change in volume level in year 1899.}
\label{fig:exp_nile_data}
\end{figure}

\paragraph{Nile data.} The interest lies in unexpected event such as natural disasters.
This data is the annual flow volume of the Nile river at Aswan from 1871 to 1970 (100 years).
In Figure \ref{fig:exp_nile_data}, the proposed algorithm results the changepoint at the $28^{\rm th}$ position, corresponding to year 1899.
This result is consistent with \citet{jung2017bayesian}.

%==============================
\subsection{Derivation of OptSeg-SI-oc mentioned in \S 5} \label{appendix:derive_dp_selection_event}%
As our first idea of SI for optimal CPs, we developed OptSeg-SI-oc. 
However, this method inherits the drawback of current SI studies (over-conditioning).
Therefore, we have not officially published it yet.
Later, we developed novel parametric programming techniques and proposed OptSeg-SI, which is presented in this paper, to address the over-conditioning problem.
Here, we show the derivation of OptSeg-SI-oc.

The main idea behinds OptSeg-SI-oc is to characterize the conditional data space based on all steps of DP algorithm, i.e., performing inference conditional on all steps of DP.
We focus on the case when $K$ is fixed, and it is easy to extend to the case when $K$ is unknown.

\paragraph{Notation. } We denote $\cX^\prime$ as a conditional data space in OptSeg-SI-oc. 
The difference between $\cX$ in \S3.1 and $\cX^\prime$ here is that the latter is characterized with additional constraints on DP process. 
For an observed sequence $\bm x^{\rm obs} \in \RR^N$, its optimal CP vector is defined as $\bm \tau^{\rm det}$. 
For a sequence with length $n \in [N]$, a set of all possible CP vectors with dimension $k \in [K]$ is defined as $\cT_{k, n}$.
Given $\bm x \in \RR^N$, the loss of segmenting its sub-sequence $\bm x_{1:n}$ with $\bm \tau \in \cT_{k, n}$ is written as 
\begin{equation*}
	L_{k, n}(\bm x, \bm \tau) = \sum \limits_{\kappa = 1}^{k + 1} C(\bmx_{\tau_{\kappa - 1} + 1: \tau_{\kappa}}).
\end{equation*}
For a sub-sequence $\bm x_{1:n}$, the optimal loss and the optimal $k$-dimensional  CP vector are respectively written as 
\begin{align*}
	L^{\rm opt}_{k, n}(\bm x) &= \min \limits_{\bm \tau \in \cT_{k, n}} L_{k, n}(\bm x, \bm \tau) \\
	\bm T^{\rm opt}_{k, n}(\bm x) &= \argmin \limits_{\bm \tau \in \cT_{k, n}} L_{k, n}(\bm x, \bm \tau).
\end{align*}

\paragraph{Conditional data space characterization.} Since the inference is conducted conditional on all steps of DP, the conditional data space $\cX^\prime$ is written as 
\begin{equation}\label{add:eq:conditional_data_space_1}
	\cX^\prime = \left\{\bm x \in \RR^{N} \mid \bigcap \limits_{k = 1}^K \bigcap \limits_{n = k}^{N} \bm T^{\rm opt}_{k, n}(\bm x) = \bm T^{\rm opt}_{k, n}(\bm x^{\rm obs}), 
q(\bm x) = q(\bm x^{\rm obs}) \right\}.
\end{equation}

For simplicity, we denote $\bm \tau^{\rm det}_{k, n} = \bm T^{\rm opt}_{k, n}(\bm x^{\rm obs})$, the conditional data space $\cX^\prime$ can be re-written as 
\begin{align}\label{add:eq:conditional_data_space_2}
	\cX^\prime &= \left\{\bm x \in \RR^{N} \mid \bigcap \limits_{k = 1}^K \bigcap \limits_{n = k}^{N} \bm T^{\rm opt}_{k, n}(\bm x) = \bm \tau^{\rm det}_{k, n}, q(\bm x) = q(\bm x^{\rm obs}) \right\}.
\end{align}
From the second condition, the data is restricted to the line \cite{liu2018more, fithian2014optimal}. Therefore, the remaining task is to characterize the region in which $\bm x \in \RR^N$ satisfies the first condition.

For each value of $k \in [K]$ and $n \in [N]$, $\bm T^{\rm opt}_{k, n}(\bm x) = \bm \tau^{\rm det}_{k, n}$ if and only if 
\begin{align}
	\min \limits_{\bm \tau \in \cT_{k, n}} L_{k, n}(\bm x, \bm \tau) &= L_{k, n}(\bm x^{\rm obs}, \bm \tau^{\rm det}_{k, n})\\
	\Leftrightarrow \quad \quad \quad \quad  L^{\rm opt}_{k, n}(\bm x) &= L_{k, n}(\bm x^{\rm obs}, \bm \tau^{\rm det}_{k, n}). \label{add:eq:ieq_1}
\end{align}

Based on the recursive structure of DP, we have 
\begin{equation}\label{add:eq:dp}
	L^{\rm opt}_{k, n}(\bm x) = \min \limits_{m \in\{k, ..., n-1\}} \left\{ L^{\rm opt}_{k-1, m}(\bm x)  + C(\bmx_{m+1:n}) \right\}.
\end{equation}
Combining \eq{add:eq:ieq_1} and \eq{add:eq:dp}, we have 
\begin{align}\label{add:eq:ieq_2}
	 L^{\rm opt}_{k-1, m}(\bm x)  + C(\bmx_{m+1:n}) \ge L_{k, n}(\bm x^{\rm obs}, \bm \tau^{\rm det}_{k, n}),
\end{align}
for $m \in\{k, ..., n-1\}$.
Since the cost function is in the quadratic form, \eq{add:eq:ieq_2} can be easily written in the form of $\bm x^\top A_{k,n,m} \bm x \leq 0$, where the matrix $A_{k,n,m} \in \mathbb{R}^{N \times N}$ depends on $k$, $n$ and $m$. It suggests that the conditional data space in \eq{add:eq:conditional_data_space_1} can be finally characterized as 

\begin{equation*}
	\cX^\prime = \left\{\bm x \in \RR^{N} \mid \bigcap \limits_{k = 1}^K \bigcap \limits_{n = k}^{N} \bigcap \limits_{m = k}^{n-1} \bm x^\top A_{k,n,m} \bm x \leq 0, 
q(\bm x) = q(\bm x^{\rm obs}) \right\}.
\end{equation*}

Now that the conditional data space $\cX^\prime$ is identified, we can easily compute the truncation region and calculate $p$-value for each detected CP.

% subset of $\cX = \{\bm x \in \RR^{N} \mid \cA(\bm x) = \cA(\bm x^{\rm obs}), \bm q(\bm x) = \bm q(\bm x^{\rm obs})\}$.

\end{document}